\documentclass{article}

     \PassOptionsToPackage{numbers, compress}{natbib}

     \usepackage[preprint]{neurips_2022}

\usepackage[utf8]{inputenc} 
\usepackage[T1]{fontenc}    
\usepackage{hyperref}       
\usepackage{url}            
\usepackage{booktabs}       
\usepackage{amsfonts}       
\usepackage{nicefrac}       
\usepackage{microtype}      
\usepackage{xcolor}         
\usepackage{amsmath}
\usepackage{amsthm}
\usepackage{todonotes}
\usepackage{glossaries}
\usepackage{import}
\usepackage{paralist}
\usepackage{lscape}

\usepackage{tikz}
\usepackage{pgfplots}
 \usetikzlibrary{
    angles,
 	arrows,
 	arrows.meta,
 	shapes,
 	shapes.symbols,
 	shapes.callouts,
 	shapes.geometric,
 	arrows,
 	matrix,
 	backgrounds,
 	positioning,
 	plotmarks,
 	calc,
 	patterns,
 	matrix,
 	decorations.pathreplacing,
 	decorations.pathmorphing,
 	decorations.text,
 	decorations.shapes,
 	decorations.fractals,
 	decorations.markings,
 	spy,
    quotes
 }
\usepgfplotslibrary{fillbetween}
\usepgfplotslibrary{statistics}
\usepgfplotslibrary{groupplots}
\usepgfplotslibrary{units}

\pgfplotsset{
  /pgfplots/confidence box/.style 2 args={
    legend image code/.code={
        \definecolor{steelblue31119180}{RGB}{31,119,180}
        \draw[steelblue31119180,no markers, fill=steelblue31119180, opacity=0.5]
        plot coordinates {
        (-0.1cm,-0.1cm)
        (-0.1cm,0.2cm)
        (0.5cm,0.2cm)
        (0.5cm,-0.1cm)
        (-0.1cm,-0.1cm)
      }
      node[rectangle]{};
    }
  }
}
 \makeatletter
 \pgfdeclareradialshading[tikz@ball]{water}{\pgfpoint{-0.15cm}{0.4cm}}{%
 	rgb(0cm)=(1,1,1);
 	color(0.35cm)=(tikz@ball!35!white); 
 	color(0.75cm)=(tikz@ball!80!white); 
 	rgb(1cm)=(1,1,1)
 }
 \tikzoption{water color}{\pgfutil@colorlet{tikz@ball}{#1}\def\tikz@shading{water}\tikz@addmode{\tikz@mode@shadetrue}}
 \def\tikz@falsetext{false}
 \tikzset{
 	shade/.code={
 		\edef\tikz@temp{#1}%
 		\ifx\tikz@temp\tikz@falsetext%
 		\tikz@addmode{\tikz@mode@shadefalse}%
 		\else%
 		\tikz@addmode{\tikz@mode@shadetrue}%
 		\fi
 	}
 }
 \makeatother
 \usepackage{pgfplots, pgfplotstable}
 \pgfplotsset{compat=1.8}
 \pgfplotstableset{
 	create on use/x/.style={
 		create col/expr={
 			\pgfplotstablerow/201*2-1
 		}
 	},
 	create on use/y/.style={
 		create col/expr accum={
 			\pgfmathaccuma+(2/201)*(abs(\pgfmathaccuma^2)+abs(\thisrow{x}^2)-1)
 		}{0.6}
 	}
 }
 \pgfplotstablenew{201}\loadedtable
 \tikzset{
 	invisible/.style={opacity=0,text opacity=0},
 	visible on/.style={alt=#1{}{invisible}},
 	alt/.code args={<#1>#2#3}{%
 		\alt<#1>{\pgfkeysalso{#2}}{\pgfkeysalso{#3}} 
 	},
 }
 \tikzset{
 	background shade/.style={#1},
 	background shade/.default={shade=false},
 	shade on/.style={alt=#1{}{background shade}},
 }
 \tikzset{
 	water hot particle/.style={
 		circle,
 		inner sep=2pt, 
 		background shade={shading=water,water color=cyan!60!black}   
 	},  
 }

 \title{Constraining Gaussian Processes to Systems of Linear Ordinary Differential Equations}

\DeclareMathOperator{\sol}{sol}
\DeclareMathOperator{\kSE}{k_{\text{SE}}}
\DeclareMathOperator{\GP}{\mathcal{GP}}

\newcommand{\R}{{\mathbb{R}}}
\newcommand{\F}{{\mathcal{F}}}

\newtheorem{theorem}{Theorem}
\newtheorem{lemma}{Lemma}

\newtheoremstyle{named}{}{}{\itshape}{}{\bfseries}{.}{.5em}{\thmnote{#3}}
\theoremstyle{named}
\newtheorem*{namedtheorem}{Theorem}

\theoremstyle{definition}
\newtheorem{example}{Example}

\newcommand{\expnumber}[2]{{#1}\mathrm{e}{#2}}
\author{%
  Andreas Besginow \\
  Department of Electrical Engineering and Computer Science\\
  OWL University of Applied Sciences and Arts\\
  Lemgo, Germany \\
  Institute industrial IT Lemgo, Germany\\
  \texttt{andreas.besginow@th-owl.de} \\
  \And
  Markus Lange-Hegermann \\
  Department of Electrical Engineering and Computer Science\\
  OWL University of Applied Sciences and Arts\\
  Lemgo, Germany \\
  Institute industrial IT Lemgo, Germany\\
  \texttt{markus.lange-hegermann@th-owl.de} \\
 }

\begin{document}

\maketitle

\begin{abstract}
    Data in many applications follows systems of Ordinary Differential Equations (ODEs).
    This paper presents a novel algorithmic and symbolic construction for covariance functions of Gaussian Processes (GPs) with realizations strictly following a system of linear homogeneous ODEs with constant coefficients, which we call LODE-GPs.
    Introducing this strong inductive bias into a GP improves modelling of such data.
    Using smith normal form algorithms, a symbolic technique, we overcome two current restrictions in the state of the art: (1) the need for certain uniqueness conditions in the set of solutions, typically assumed in classical ODE solvers and their probabilistic counterparts, and (2) the restriction to controllable systems, typically assumed when encoding differential equations in covariance functions.
    We show the effectiveness of LODE-GPs in a number of experiments, for example learning physically interpretable parameters by maximizing the likelihood.
\end{abstract}

\newacronym{dl}{DL}{Deep Learning}
\newacronym{ev}{EV}{Eigenvalue}
\newacronym[plural=GPs,firstplural=Gaussian Processes (GPs)]{gp}{GP}{Gaussian Process}
\newacronym{lmc}{LMC}{linear model of coregionalization}
\newacronym{lfm}{LFM}{Latent Force Model}
\newacronym{lodegp}{LODE-GP}{}\glsunset{lodegp}
\newacronym{ml}{ML}{Machine Learning}
\newacronym[plural=MOGPs,firstplural=Multi-Output Gaussian Processes (MOGPs)]{mogp}{MOGP}{Multi-Output Gaussian Process}
\newacronym{nn}{NN}{Neural Network}
\newacronym{pml}{PML}{Perfectly Matched Layers}
\newacronym{rmse}{RMSE}{}\glsunset{rmse}
\newacronym{se}{SE}{Squared Exponential}
\newacronym{snf}{SNF}{Smith Normal Form}
\newacronym{ode}{ODE}{Ordinary Differential Equation}
\glsdisablehyper

\section{Introduction}

Many real world tasks have underlying dynamic behavior, for example chemical reactions \cite{goeke2012computing}, systems in bioprocess engineering \cite{hernandez2022designing}, or population dynamics \cite{wangersky1978lotka}.
One currently widely debated application are compartmental models in epidemiology \cite{fanelli2020analysis,kumar2020covid,schmidt2021probabilistic}.
Many such systems are linear or can be decently linearized, such as in control theory \cite{zerz}, biology \cite{de2002inferring}, process engineering \cite[\textsection9]{adkins2012linear}, or engines \cite{bertin2000thermal}.
Including prior knowledge in the form of differential equations benefits the model fit and enhances interpretability for a model.
Hence, modelling differential equations in \gls{ml} has therefore been the focus of much research in both \glspl{gp} (e.g.\ \cite{langehegermann2018algorithmic, jidling2017linearly,sarkka2011linear, gonccalves2021machine, ulaganathan2016performance}) and \gls{dl} (e.g.\ \cite{raissi2019physics, lagaris2000neural, milligen1995neural, drygala2022generative}).

The class of probabilistic \gls{ode} solvers allow to estimate the solution of \gls{ode} initial value problems through approximations, often based on e.g.\ Runge-Kutta methods or Kalman filters \cite{schmidt2021probabilistic,schober2019probabilistic, schober2014probabilistic, bosch2021calibrated}, and thereby do not strictly guarantee to yield solutions of the \gls{ode}.
This class of algorithms is commonly used to solve non-linear \glspl{ode}, but typically require that systems are well-posed, in particular the solution needs to be unique once a finite number of initial conditions is known.
While this second limitation is typically irrelevant in physical or biological systems, it is strongly relevant in systems in engineering, such as in control systems with their freely choosable inputs.

With early works like \cite{graepel2003solving}, the introduction of derivatives as linear operators in \glspl{gp} became prominent for partial differential equations.
This inspired several works that encode differential equations in the covariance structure of \glspl{gp} \cite{jidling2017linearly, sarkka2011linear, gonccalves2021machine, ulaganathan2016performance, raissi2017machine} to introduce a strong inductive bias into \glspl{gp}.
This motivated \cite{langehegermann2018algorithmic} to make these approaches algorithmic, build a mathematical foundation, and show the hidden assumption in previous works of being limited to controllable systems.

This paper overcomes the necessity of approximations%
, the restriction to systems where initial conditions lead to unique solutions, and the restriction to controllable systems.
For that purpose, we algorithmically construct \glspl{lodegp} (Theorem \ref{theorem:gp_model_from_snf}), a novel class of \glspl{gp} whose realizations strictly follow a given system of homogenuous
linear \glspl{ode} with constant coefficients.

Our construction of \glspl{lodegp} starts by describing our system of \glspl{ode} via a so-called operator matrix $A$.
Calculating its \gls{snf} gives us a decoupled representation of the system via a diagonal operator matrix $D$, at the cost of going through two invertible base change matrices $U, V$  with the relationship $U\cdot A\cdot V = D$.
We can easily construct a multi-output \gls{gp} $g=\GP(0,k)$ (cf.\ Lemma~\ref{lemma:base_cov_table}) with as many outputs as the system channels and whose covariance function encodes the decoupled system.
Then, applying the base change matrix $V$ to $g$ yields the pushforward \gls{gp} $V_*g=\GP(0,VkV')$ whose covariance function encodes the original system, i.e.\ the \gls{lodegp}.
We refer to Theorem~\ref{theorem:gp_model_from_snf} and its algorithmic proof for details.

Our paper makes the following contributions:
\begin{enumerate}
    \item It develops and proves an algorithmic and symbolic construction of \glspl{lodegp}, \glspl{gp} with the strong inductive bias that their realizations strictly satisfy \emph{any} given system of homogeneous linear \glspl{ode} with constant coefficients (see Theorem~\ref{theorem:gp_model_from_snf}).
    \item It demonstrates that the constructed \glspl{gp} are numerically stable and robust (see Section~\ref{sec:evaluation}).
    \item It automatically includes \gls{ode} system parameters as \gls{gp} parameters, learns them during the training process of the \gls{gp}, and allows an interpretation of the data (see Subsection~\ref{subsection:heating}).
    \item It provides a free and public implementation based on the \gls{gp} library GPyTorch\footnote{github link will be added after acceptance of the paper; code is in the supplmentary material.} \cite{gardner2018gpytorch}. 
\end{enumerate}

We successfully test our approach on controllable and non-controllable systems of differential equations, with and without system parameters.
The \gls{lodegp} strictly satisfies the \glspl{ode} and outperforms \glspl{gp} by several magnitudes in its precision, additionally it correctly reconstructs the system parameters used to generate the data with a small relative error, even despite noise.
We also discuss how this strict behavior influences results for metrics like the \gls{rmse} with e.g.\ Figure~\ref{fig:bipendulum_adam_mean_comparison}.

\section{Related Work}
Using physical information in \gls{ml} systems became a central research topic over recent years in both \gls{dl} and \glspl{gp}.
With the \gls{gp} research in two important categories: (1) Works that introduce derivative information as inductive bias into the \gls{gp} \cite{jidling2017linearly, langehegermann2018algorithmic, langehegermann2021linearly, graepel2003solving, sarkka2011linear, gonccalves2021machine} and (2) works that train \glspl{gp} without inductive bias on physical data, with specially selected standard kernels, as an approximate model for additional processing steps
\cite{wahlstroem2013modeling,  ulaganathan2016performance, solin2018modeling, klenske2015gaussian, bilionis2013multi, calderhead2009accelerating, chai2008multi}.
Of all the works on \glspl{gp}, \cite{langehegermann2018algorithmic, langehegermann2021linearly} are the closest to us, as they also provide general algorithms to introduce inductive bias into \glspl{gp}, based on Gröbner bases.
These approaches are only applicable to controllable systems.
By basing our algorithm on the \gls{snf} instead of Gröbner bases, we overcome this limitation to controllable systems, but are restricted to \emph{\glspl{ode}}.

Probabilistic \gls{ode} solvers like \cite{schmidt2021probabilistic,kramer2021linear,tronarp2021bayesian,bosch2021calibrated,marco2015automatic,schober2014probabilistic, schober2019probabilistic,biegler1986nonlinear,calderhead2009accelerating} are able to include inductive bias of \glspl{ode} but sometimes require a number of comparably complex calculations and some approximations.
We avoid any approximation through calculation of the \gls{snf} and the application of the pushforward with a linear differential operator.
Hence, our work is limited to \emph{linear} \glspl{ode}, whereas probabilistic \gls{ode} solvers are able to work with non-linear \glspl{ode}.

For \gls{gp} priors for decoupled \gls{ode} systems with constant coefficients and right hand side functions see \cite{alvarez2009latent}, which consideres the right hand side functions as latent, hence these models are called \gls{lfm}.
A \gls{gp} prior on these latent forces and is pushed forward through differential operators and Green's operator.
We empirically compare \glspl{lodegp} to \glspl{lfm} in the appendix.

Of the other works that introduce inductive bias, \cite{graepel2003solving} and \cite{sarkka2011linear} are the earliest works and already discuss a general formulation of differential equations by viewing derivatives as linear operators that can be applied to \glspl{gp}.
Subsequent works target specific \glspl{ode} and create covariance functions with inductive bias \cite{gonccalves2021machine, ulaganathan2016performance, cross2021physics}.
They can be considered special cases of our approach with a specific manually constructed pushforward.
Finally, others constructed covariance functions for \emph{partial} differential equations from standard covariance functions with e.g.\ curl-free behaviour \cite{wahlstroem2013modeling} to model physical behaviour \cite{wahlstroem2013modeling, ulaganathan2016performance, solin2018modeling, klenske2015gaussian} or use the \gls{gp} as a surrogate model to learn physical problems like the inverse pole problem \cite{chai2008multi} or a variety of applications \cite{bilionis2013multi}.

Most \gls{dl} models explore physically informed models through additional loss terms that punish solutions that deviate from the system \cite{milligen1995neural, lagaris2000neural, raissi2019physics}, or in the case of \cite{drygala2022generative} show the equivalence of the loss to their assumed physical behavior.
This is often combined with different modifications to the networks like additional features \cite{drygala2022generative} or specific network structures \cite{lagaris2000neural}.

\section{Background}\label{sec:relatedWork}

\subsection{Gaussian Processes}\label{sec:GPs}
A \gls{gp} $g = \mathcal{GP}(\mu, k)$ defines a probability distribution over the space of functions $\mathbb{R}^d \to \mathbb{R}^\ell$, such that the outputs $g(x_i)$ at any set of $x_i \in \mathbb{R}^d$ are jointly Gaussian \cite{rasmussen2006gaussian}.
Such a (multi-output) \gls{gp} is defined by its mean function (often set to zero)
$$
\mu : \mathbb{R}^d \to \mathbb{R}^\ell : x \mapsto \mathbb{E}(g(x))
$$
and its (multi-output) positive semi-definite covariance function (also called kernel)
$$
k : \mathbb{R}^d \times \mathbb{R}^d \to \mathbb{R}^\ell_{\succeq 0} : (x, x') \mapsto \mathbb{E}((g(x) - \mu(x))(g(x') - \mu(x'))^T).
$$
A popular kernel is the \gls{se} kernel $k_{SE}(x, x') = \sigma^2 \exp\left(-\frac{(x-x')^2}{2\ell^2}\right)$, with its signal variance $\sigma$ and lengthscale $\ell$.
It models smooth data very well and is thus usable for many datasets~\cite[p. 83]{rasmussen2006gaussian}, as its realizations are dense in the space of smooth, i.e.\ infinitely differentiable, functions $C^\infty(\R,\R)$ \cite[Prop.~1]{langehegermann2022boundary}.

Manipulating existing GPs allows to introduce indutive biases in various ways \cite{duvenaud2014automatic,duvenaud2013structure,snelson2003warped,calandra2016manifold,thewes2015advanced}.
One important example is the application of matrices of linear operators to a \gls{gp}.
Formally, this is the pushforward operation of an operator matrix $B$ on the \gls{gp} $g$ as $B_{*}g = \mathcal{GP}(B\mu(x), Bk(x, x')(B')^T)$ \cite{berlinet2011reproducing}, where $B'$ denotes the operation of $B$ on the second argument of $k(x, x')$ \cite[Lemma 2.2]{langehegermann2021linearly}.
These operators can, for example, be differential operators. 
The matrix $B$ induces the strong bias such that all realizations lie in the image of $B$.
This pushforward is typical for applications in differential equations \cite{jidling2017linearly, langehegermann2018algorithmic, langehegermann2021linearly, graepel2003solving, sarkka2011linear} or geometry \cite{hutchinson2021vector}.

\begin{example}
    \glspl{gp} can be constrained to realizations satisfying a system of linear equations given by a matrix $A$, e.g.\ 
    $A =\begin{bmatrix} 2 & -3 \end{bmatrix}$
\begin{align*}
    \sol_\F(A) :\!&= \left\{\begin{bmatrix} f_1(x) & f_2(x)\end{bmatrix}^T\in\F^{2\times1} \,\middle|\, A \cdot\begin{bmatrix} f_1(x) & f_2(x)\end{bmatrix}^T =0\right\}.
\end{align*}
with $\F=C^\infty(\R,\R)$ the space of smooth input functions.
The matrix $B =\begin{bmatrix} 3 \\ 2 \end{bmatrix}$ is maximal in that it solves $A\cdot B=0$ and hence $\sol_\F(A) = B\cdot\F = \left\{ B\cdot f(x) \,\middle|\, f(x)\in\F \right\}$.
Taking a GP prior $g=\GP(0,k)$ for $f(x)\in\F$ and applying the pushforward then yields a new, constrained GP prior
\begin{align*}
    B_*g = \GP(0,Bk(B')^T) = \GP\left(
       \begin{bmatrix} 0 \\ 0 \end{bmatrix}, 
       \begin{bmatrix}
        9\cdot \kSE & 6\cdot \kSE \\
        6\cdot \kSE & 4\cdot \kSE
\end{bmatrix}\right)
\end{align*}
for $B\cdot f(x)\in\sol_\F(A)$, whose realizations are guaranteed to lie in $\sol_\F(A)$.

This example is similar to our use case of systems of differential equations for \glspl{lodegp}, where we just replace the matrix $B$ of numbers with a suitable matrix of differential operators.
\end{example}

\subsection{Smith normal form}
The \gls{snf} \cite{smith1862systems, newman1997smith} is a normal form for a matrix $A\in \R[x]^{m\times n}$ over the polynomial ring $\R[x]$, such that $U \cdot A \cdot V = D$.
Here, $D\in \R[x]^{m\times n}$ is a (not necessarily square) diagonal matrix of same size as $A$ and the base change matrices $U \in \R[x]^{m\times m}$ and $V \in \R[x]^{n\times n}$ are invertible square matrices, i.e.\  $\operatorname{det}(U), \operatorname{det}(V) \in \R \setminus\{0\}$ \cite{gollmann2017lineare, cluzeau2012symbolic}.
Algorithms to construct the \gls{snf} are implemented in many computer algebra systems such as Matlab \cite{MATLAB}, Maple \cite{maple}, or SageMath \cite{sagemath} (based on PARI \cite{PARI2}) which is a free and open source python library for computer algebra.
Intuitively, the \gls{snf} can be compared to the eigendecomposition of a $\mathbb{R}^{n\times n}$ matrix, which---if it exists---produces a square matrix of eigenvectors $W$ and the diagonal matrix of eigenvalues $\Lambda$.
This analogy is lacking since e.g.\ the \gls{snf} produces two independent base change operator matrices $V$ resp.\ $U$  instead of a single eigenvector matrix $W$ resp.\ its inverse $W^{-1}$.
Still, the matrix $D$ decouples and thereby simplifies a system given by the input matrix $A$, as does the diagonal eigenvalue matrix $\Lambda$ \cite{robertz2006formal, cluzeau2012symbolic}. 

Since the \gls{snf} exists for matrices over polynomial rings over any field, we can also compute it over a polynomial ring over the function field $\R(a_1,\ldots,a_k)$.
Hence, we can model differential equations containing parameters $a_1,\ldots,a_k$.
The heating example in Subsection~\ref{subsection:heating} demonstrates how such parameters are being used in the \gls{snf} and subsequently learned from data.

\begin{example}\label{example:bipendulum_init}
A bipendulum acts as our running example.
It consists of a rod with a pendulum attached to each end of the rod, see Figure \ref{fig:bipendulum_visualization}.
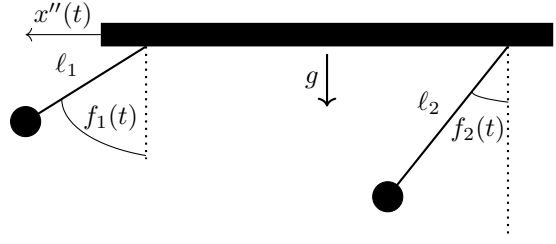
\begin{figure}
    \center
    \begin{tikzpicture}
    \node[rectangle,
        draw = black,
        fill = black,
        minimum width = 6cm,
        minimum height = 0.3cm] (r) at (0,0) {}; 
    \draw [->] (r.west) -- ($(r.west) + (-1, 0)$);
    \node (x) at ($($(r.west) + (0,0.5)$)!0.5!($(r.west) + (-1, 0)$)$) {$x''(t)$};
    \coordinate[] (right_rope_end) at ($(r.south east) + (-2.2,-2)$);
    \coordinate[] (right_rope_start) at ($(r.south west)!0.9!(r.south east)$);
    \coordinate[] (left_rope_end) at ($(r.south west) + (-1,-1)$);
    \coordinate[] (left_rope_start) at ($(r.south west)!0.1!(r.south east)$);
    \draw[thick] (left_rope_start) -- (left_rope_end);
    \filldraw (left_rope_end) circle [radius=0.2];
    \node[inner sep=0](ell1) at ($($(left_rope_start) + (-0.5,0)$)!0.5!($(left_rope_end) + (0, 0.5)$)$) {$\ell_1$};
    \draw[thick] (right_rope_start) -- (right_rope_end);
    \filldraw (right_rope_end) circle [radius=0.2];
    \node[inner sep=0](ell2) at ($($(right_rope_start) + (-0.5,0)$)!0.5!($(right_rope_end) + (0, 0.5)$)$) {$\ell_2$};
    \coordinate (gravity_start) at ($(r.south) + (0,-0.1)$);
    \coordinate (gravity_end) at ($(r.south) + (0,-0.8)$);
    \draw[thick, ->] (gravity_start) -- (gravity_end);
    \node[inner sep=0](gravity_text) at ($($(gravity_start)+(-0.2, 0)$)!0.5!($(gravity_end)+(-0.2, 0)$)$) {$g$};
    \draw[thick, dotted] let \p1 =(left_rope_start), \p2 = (left_rope_end) in (left_rope_start) -- ($(\x1, \y2) + (0, -0.5)$);
    \draw[thick, dotted] let \p1 =(right_rope_start), \p2 = (right_rope_end) in (right_rope_start) -- ($(\x1, \y2) + (0, -0.5)$);
    \path let \p1 =(left_rope_start), \p2 = (left_rope_end) in coordinate (left_vertical_axis_end) at ($(\x1, \y2) + (0, -0.5)$);
    \draw ($(left_rope_end)!0.3!(left_rope_start)$) arc [start angle=0, delta angle=-70, x radius=-1.7cm, y radius=0.8cm];
    \node[] (left_angle_text) at ($(left_rope_end)+(1.13, 0.05)$) {$f_1(t)$};
    \path let \p1 =(right_rope_start), \p2 = (right_rope_end) in coordinate (right_vertical_axis_end) at ($(\x1, \y2) + (0, -0.5)$);
    \draw ($(right_rope_end)!0.7!(right_rope_start)$) arc [start angle=0, delta angle=-45, x radius=-1.65cm, y radius=0.2cm];
    \node[] (right_angle_text) at ($(right_rope_end)+(1.20, 0.89)$) {$f_2(t)$};
    
    \end{tikzpicture}
    \label{fig:bipendulum_visualization}
    \caption{A visualization of the bipendulum with its components (rope lengths $\ell_1, \ell_2$) and states (angles $f_1(t), f_2(t)$, rod acceleration $x''(t)$).}
\end{figure}
The linearized equations of an idealized bipendulum:
\begin{align}\label{eq:bipendel_system_equations}
   \begin{bmatrix}
    x''(t) + \ell_1f_1''(t) + gf_1(t) \\
    x''(t) + \ell_2f_2''(t) + gf_2(t) 
\end{bmatrix} = \mathbf{0} = 
\underbrace{
    \begin{bmatrix}
        \partial_t^2 + \frac{g}{\ell_1} & 0 & -\frac{1}{\ell_1} \\
        0 & \partial_t^2 + \frac{g}{\ell_2} & -\frac{1}{\ell_2}
    \end{bmatrix}}_A
    \cdot
    \begin{bmatrix}
        f_1(t) \\
        f_2(t) \\
        u(t)
    \end{bmatrix}
\end{align}
with the operator matrix $A$, $\ell_i > 0$ the length of the $i$-th pendulum, $g = 9.81$ the gravitational constant, $u(t)=-x''(t)$ the rod's acceleration, and $f_i(t), i=1,2$ the angle between the $i$-th pendulum and the vertical axis.
The \gls{snf} of $A$ reveals its properties, specifically when $\ell_1 = \ell_2$ and $\ell_1 \neq \ell_2$.
An algorithm calculates the \gls{snf} $D$ of $A$ and the base change matrices $U,V$ as follows:
\begin{align*}
    &\text{case } \ell_1=1 \neq \ell_2=2\text{:}\\
    &\underbrace{
    \begin{bmatrix}
    1 & 0 \\
    -\frac{1}{2} & 1
    \end{bmatrix}
    }_U
    \underbrace{
    \begin{bmatrix}
        \partial_t^2 + g & 0 & -1 \\
        0 & \partial_t^2 + \frac{g}{2} & -\frac{1}{2}
    \end{bmatrix}
    }_A
    \underbrace{
    \begin{bmatrix}
            0 & -\frac{4}{g} & \frac{2\partial_t^{2}+g}{2} \\
            0 & -\frac{2}{g} & \frac{\partial_t^{2}+g}{2} \\
        -1 & -\frac{4\partial_t^{2}+4g}{g}  & (\partial_t^2+\frac{g}{2})(\partial_t^2+g)
    \end{bmatrix}
    }_V
    =
    \underbrace{
    \begin{bmatrix}
    1 & 0 & 0 \\
    0 & 1 & 0
    \end{bmatrix}
    }_D\\
    &\text{case } l_1 = l_2 = 1\text{:}\\
    &\underbrace{
    \begin{bmatrix}
    1 & 0 \\
    -1 & 1
    \end{bmatrix}
    }_U
    \underbrace{
    \begin{bmatrix}
        \partial_t^2+g & 0 & -1 \\
        0 & \partial_t^2 + g & -1
    \end{bmatrix}
    }_A
    \underbrace{
    \begin{bmatrix}
    0 & 0 & 1 \\
    0 & 1 & 1 \\
    -1 & 0 & \partial_t^{2} + g 
    \end{bmatrix}
    }_V
    =
    \underbrace{
    \begin{bmatrix}
    1 & 0 & 0 \\
    0 & \partial_t^{2} + g & 0 \\
    \end{bmatrix}
    }_D
\end{align*}

In case of unequal rope length, the controllability is intuitive when imagining that, by accelerating the rod at the right time, the two ropes can reach every combination of valid positions.
Formally, this follows that $D\cdot\begin{bmatrix}h_1(t)&h_2(t)&h_3(t)\end{bmatrix}^T=0$ if and only if $h_1(t)=h_2(t)=0$ and $h_3(t)$ is an arbitrarily choosable output.
In case of equal rope length, the system is not controllable, indicated by the diagonal entry $\partial_t^2 + g$ of $D$
\cite[p. 154]{oberst1990multidimensional}. 
\end{example}

\section{Constructing a GP for differential equations}\label{sec:method}

We introduce a polynomial time algorithm to construct \glspl{lodegp}, a class of \glspl{gp} with realizations dense in the space of solutions of linear \glspl{ode} with constant coefficients.
This ensures that the \gls{lodegp} is able to produce all possible solutions to the \glspl{ode} and nothing but solutions for the \glspl{ode}.
All this is guaranteed to be strictly accurate, since we make no approximations.

Consider a system of linear homogenous ordinary differential equations with constant coefficients
\begin{equation}\label{eq:ode_base_eq}
    A\cdot \mathbf{f}(t) = 0 
\end{equation}
with operator matrix $A \in \mathbb{R}[\partial_t]^{m\times n}$ determining the relationship between the smooth functions $f_i(t) \in C^\infty(\R, \R)$ of $\mathbf{f}(t)=\left(\begin{matrix}f_1(t) & \hdots & f_n(t)\end{matrix}\right)^T$.
For such systems our main result holds.
\begin{theorem}{(\glspl{lodegp})}\label{theorem:gp_model_from_snf}
    For every 
    system
    as in Equation~\eqref{eq:ode_base_eq} there exists a \gls{gp} $g$, such that the set of realizations of $g$ is dense in the set of solutions of $A\cdot\mathbf{f}(t)=0$.
\end{theorem}

The following lemma play a crucial role in the proof of this theorem by constructing base covariance functions for a system as in Equation~\eqref{eq:ode_base_eq} that is \emph{decoupled} (via the \gls{snf}) into scalar equations.
The proof in Appendix~\ref{app:proof_kernel_construction} makes use of the solution space being decomposed, in particular is deal wih finite dimensional vector space like Bayesian linear regression.
\begin{lemma}\label{lemma:base_cov_table}
    Covariance functions for solutions of the scalar linear differential equations $d\cdot f = 0$ with constant coefficients, i.e.\ $d\in\R[\partial_t]$, are given by Table~\ref{tab:base_cov_construction_paper} for $d$ is primary, i.e.\ a power of an irreducible real polynomial.
    In the case of a non-primary $d$, each primary factor $d_i$ of $d=\prod_{i=0}^{\ell-1}d_i$ is first translated to its respective covariance function $k_i$ separately before they are added up to give the full covariance function $k(t_1, t_2) = \sum_{i=0}^{\ell-1} k_i(t_1, t_2)$. 
    \begin{table}[h]
    \caption{Primary operators $d$ and their corresponding covariance function $k(t_1, t_2)$.}
        \centering
    \begin{tabular}{cc}
    	\toprule
        $d$ & $k(t_1, t_2)$\\
        \midrule
        $1$ & $0$ \\
        $(\partial_t -a)^j$ & $\left(\sum_{i=0}^{j-1} t_1^it_2^i\right)\cdot \exp(a\cdot (t_1+ t_2))$ \\
        $((\partial_t -a - ib)(\partial_t -a + ib))^j$ & $\left(\sum_{i=0}^{j-1} t_1^it_2^i\right)\cdot \exp(a\cdot (t_1 + t_2))\cdot\cos(b\cdot( t_1-t_2))$ \\
        $0$ & $\exp(-\frac{1}{2}(t_1-t_2)^2)$ \\
        \bottomrule
    \end{tabular}
    \label{tab:base_cov_construction_paper}
    \end{table}
\end{lemma}

For the case of single (i.e.\ $j=1$) zeroes, the sum in the covariance function simplifies to a factor $1$.

\begin{example}
    The \gls{ode} $d=\partial_t -1$ has solutions of the form $a\cdot \exp(t)$ for $a\in \mathbb{R}$. This one-dimensional solution space is described by the covariance function $k(t_1, t_2)=\exp(t_1)\cdot\exp(t_2)=\exp(t_1+t_2)$.
\end{example}

\begin{proof}[Proof of Theorem \ref{theorem:gp_model_from_snf}]
    We begin with a neutral multiplication of $A\cdot \mathbf{f}=0$ with $V\cdot V^{-1}$ and a left multiplication by $U$ to decouple the system using the \gls{snf} $D=U\cdot A\cdot V$ (cf.\ Equation \eqref{eq:ode_base_eq}):
    \begin{equation}
    \begin{aligned}\label{eq:DiffEq_Lf_eq_0}
        && U\cdot A\cdot V\cdot V^{-1}\cdot \mathbf{f} &= 0\\
        \Leftrightarrow && D\cdot V^{-1}\cdot \mathbf{f} &= 0\\
        \Leftrightarrow && D\cdot \mathbf{p} &= 0 \\
        \Leftrightarrow && \bigwedge_{i=1}^{\min(n,m)} D_{i,i}\cdot\mathbf{p}_i&=0 \quad \wedge \quad \bigwedge_{i=\min(n,m)+1}^n 0\cdot \mathbf{p}_i=0
    \end{aligned}
    \end{equation}
    for decoupled latent vector $\mathbf{p} = V^{-1}\mathbf{f}$ of functions.

    We assume the \gls{gp}-prior $h \sim \mathcal{GP}(\mathbf{0}, k)$ for this vector $\mathbf{p}$ via a \gls{gp}, where $k$ is a multi-output diagonal covariance function such that each diagonal entry of $k$ is given by Lemma~\ref{lemma:base_cov_table}.
    By this Lemma, the set of realizations of this prior $h$ is dense in the set of solutions of $D\cdot \mathbf{p} = 0$.

    The pushforward of $h$ with $V$ yields a \gls{gp} $g$ that can learn the original system of \glspl{ode}.
    \begin{equation}\label{eq:GPpushforwardT}
        g\sim V_*h = \mathcal{GP}(\mathbf{0}, V\cdot k \cdot V')
    \end{equation}
    with $V' = V^T$ the operation applied on the second entry of the kernel (i.e.\ $t_2$) using \cite[Lemma~2.2]{langehegermann2021linearly}.
    As $V$ is invertible, the set of realizations of $g$ is dense in the set of solutions of $A\cdot \mathbf{f}=0$.
\end{proof}

This proof shows that diagonalizing the system to $D\cdot\mathbf{p}=0$ decouples the components of $\mathbf{p}$ to expose the intrinsic behavior of the system.
We can also easily interpret the controllability of a system in this decoupled form.
A zero column (or $0\cdot \mathbf{p}_i=0$) stands for a freely choosable function in $\mathbf{p}$, i.e.\ an output of the system.
A one in a column (or $1\cdot \mathbf{p}_i=0$) stands for something we have no choice in, e.g.\ an input or a state that is fixed once the desired output was chosen.
Any other entry leads to sinusoidal-exponential behavior, which is uncontrollable in the sense that it cannot be influenced.
This behaviour is classic in systems of differential equations, as most of them can be split into a controllable and uncontrollable solution space, which can be clearly seen in the entries of $D$.
Summing up, a system is controllable iff all diagonal entries of $D$ are either zero or one.

This paper improves upon \cite{langehegermann2018algorithmic} for two reasons:
first, it decouples the uncontrollable part of a system via the \gls{snf} as a direct summand and, second, it constructs a covariance function for this uncontrollable summand.
This is generally impossible for partial differential equations due to \cite{BaPurity_MTNS10,QGrade}.

The runtime of the construction of \glspl{lodegp} in Theorem~\ref{theorem:gp_model_from_snf} is dominated by computing the \gls{snf}.
This has a deterministic polynomial complexity in the size of the (usually small) operator matrices $A$ \cite{villard1993computation,villard1995generalized,labhalla1996algorithmes,wilkening2011local} with quicker probabilistic Las Vegas algorithms \cite{storjohann1997fast}, and it can be parallelized
\cite{villard1994fast,villard1997fast}.
For our examples, the \gls{snf} terminates instantaniously.
The application of the base change matrix $V$ is cubic in the (usually small) size of matrices, assuming constant time to apply operators to the covariance functions from Lemma~\ref{lemma:base_cov_table}.
Once the covariance function is constructed, the complexity of $\mathcal{O}(n^3)$ for \glspl{gp} applies, where $n$ is the (potentially big) number of data points.

\begin{example}
    We continue Example~\ref{example:bipendulum_init} and consider the \gls{snf} of $A$ for $l_1 \neq l_2$.
    Since the three diagonal entries of the matrix $D$ are $1, 1, 0$ we conclude that latent kernel $k$ has diagonal entries $0, 0, \kSE$ with Lemma~\ref{lemma:base_cov_table}.
    Now the pushforward creates the \gls{lodegp}, which requires the following application of two operator matrices; we simplify by removing irrelevant zeroes of the latent covariance.
    \begin{align*}
    VkV' &=
    \begin{bmatrix}
        \frac{2\partial_{t_1}^{2}+g}{2} \\
        \frac{\partial_{t_1}^{2}+g}{2} \\
        (\partial_{t_1}^2+\frac{g}{2})(\partial_{t_1}^2+g)
    \end{bmatrix}
    \cdot
    \begin{bmatrix}
    \kSE 
    \end{bmatrix}
    \cdot
    \begin{bmatrix}
        \frac{2\partial_{t_2}^{2}+g}{2}& 
        \frac{\partial_{t_2}^{2}+g}{2}& 
        (\partial_{t_2}^2+\frac{g}{2})(\partial_{t_2}^2+g)
    \end{bmatrix}
    \end{align*}
    Applying these two operators to the kernel $\kSE$ results in the covariance function for the bipendulum.
\end{example}

\section{Experimental evaluation}\label{sec:evaluation}

\begin{table}[h]
    \caption{The median results and standard dev. for training and evaluation \gls{rmse}, loss value (negative marginal log likelihood per datapoint) and the mean \gls{ode} satisfaction error, over 20 experiments for the \gls{gp} and \gls{lodegp} on noisy training data. The mean \gls{ode} error is the average of each \gls{ode} error for a function in a system. The \glspl{rmse} were calculated with noiseless data in the training and evaluation interval, the \gls{ode} error only with the latter, for each respective experiment. Smaller is better.}
        \centering
    \begin{tabular}{cccccc}
    	\toprule
        && training \gls{rmse} & evaluation \gls{rmse} & loss & mean \gls{ode} error\\
        \midrule
        Bipendulum &\gls{lodegp} & 0.060 $\pm$ 0.060 & 0.106 $\pm$ 0.258 & -1.638 $\pm$ 0.567 & \textbf{5.135e-07} $\pm$ 1.607e-06\\
                   &\gls{gp}     & \textbf{0.008} $\pm$ 0.030& \textbf{0.044} $\pm$ 0.0175 & \textbf{-2.072} $\pm$ 0.215 & 0.064 $\pm$ 0.023 \\
        \midrule
        Heating    &\gls{lodegp} & \textbf{0.006} $\pm$ 0.001& \textbf{0.170} $\pm$ 0.068 & -\textbf{2.089} $\pm$ 0.100 & \textbf{0.008} $\pm$ 0.007\\
                   &\gls{gp}     & 0.010 $\pm$ 0.001& 0.333 $\pm$ 0.055 & -1.628 $\pm$ 0.051 & 0.218 $\pm$ 0.070 \\
        \midrule
        Three tank &\gls{lodegp} & \textbf{0.023} $\pm$ 0.004& 0.078 $\pm$ 0.046 & -0.949 $\pm$ 0.063 & \textbf{1.360e-05} $\pm$ 4.919e-06\\
                   &\gls{gp}     & 0.028 $\pm$ 0.005& \textbf{0.058} $\pm$ 0.031 & \textbf{-0.974} $\pm$ 0.071 & 0.040 $\pm$ 0.020\\
        \bottomrule
    \end{tabular}
\label{tab:all_eval_results}
\end{table}
We demonstrate the effectiveness of \glspl{lodegp} in three examples, where we constrain a \gls{lodegp} using the system description and train it with datapoints, similar to solving an initial value problem.
The Gröbner basis approach of \cite{langehegermann2018algorithmic} yields precisely the same results for the two controllable examples and is not applicable to the uncontrollable example.
The only other method that could deal with our class of differential equations is \cite{alvarez2009latent}\footnote{Most probabilistic \gls{ode} solvers are not applicable to systems with free functions in their solution set, with the exception of \cite{schmidt2021probabilistic}, which can only estimate free functions of parameters.}, for which a comparison with the three tank system is discussed in the appendix, in the following we compare our model mainly to classic \glspl{gp}.

Our comparison includes the error in satisfying the \glspl{ode}, specified by the median error the \glspl{gp} posterior mean function has in satisfying the \glspl{ode} at evenly spread points, where we calculate derivatives through finite differences.
The \glspl{lodegp} hyperparameters (lengthscale and signal variance) are randomly initialized from a uniform distribution on $[-3,3]$ and were trained using Adam \cite{kingma2015adam}.
The 25 training datapoints are created uniformly in the intervals $[1, 6]$ and $[-5, 5]$ for the bipendulum and three tank system, resp.\ the heating system.
Training and evaluation is repeated ten times in each experiment using a GPyTorch \cite{gardner2018gpytorch} implementation of our \gls{lodegp} construction with SageMath \cite{sagemath} to symbolically calculate the \gls{snf}.
The training loss is the negative marginal log likelihood (c.f.\ \cite[Eq. (2.30)]{rasmussen2006gaussian}) which, in GPyTorch, is additionally divided by the number of training data points.
Details of the training, data generation, verification process, additional experiments, e.g.\ without noise, and further system details can be found in the appendix.

\subsection{Bipendulum - Controllable}
We continue with the controllable variant of the bipendulum from Example~\ref{example:bipendulum_init} with the rope lengths $\ell_1 = 1$ and $\ell_2 = 2$.
We create 25 points of training data from a solution of the \glspl{ode} as shown in Figure~\ref{fig:bipendulum_posterior_our}.
We add white noise to the data with standard deviation of 2\% of the maximal signal.

\begin{figure}
    \centering
    \import{}{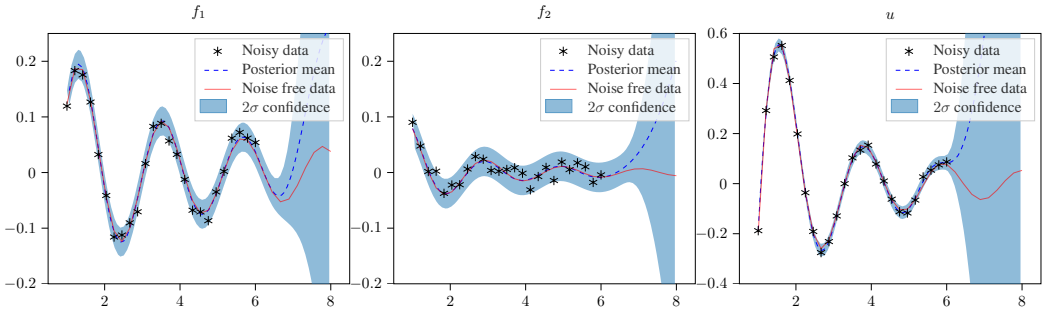}
    \caption{A posterior \gls{lodegp} trained on noisy bipendulum data (black stars), its confidence (blue transparent), posterior mean (blue dashed line) and the original noise-free function (red line). The posterior mean fits the original data very well and shows that the behavior is learned despite noise.}
    \label{fig:bipendulum_posterior_our}
\end{figure}

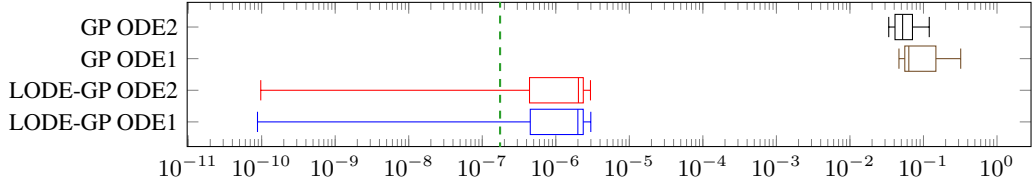
\begin{figure}
    \center
    \begin{tikzpicture}
    \definecolor{forestgreen4416044}{RGB}{44,160,44}
    \begin{axis}
    [width=12.8cm,
    height=3.5cm,
    xmode = log,        
    ytick={1,2,3,4},
    yticklabels={LODE-GP ODE1, LODE-GP ODE2, GP ODE1, GP ODE2}, 
    font=\footnotesize
    ]
    \draw [dashed, thick, forestgreen4416044] (-15.56,-100) -- (-15.56,500);
        \addplot+[
    boxplot prepared={
        median        =1.995426793316254e-06,
        upper quartile=2.3613691123780263e-06,
        lower quartile=4.4984243302757346e-07,
        upper whisker =2.993270496693754e-06,
        lower whisker =8.781876982641347e-11    
    },
    ] coordinates {};
    \addplot+[
    boxplot prepared={
        median        =2.031589538482337e-06,
        upper quartile=2.3654009184674914e-06,
        lower quartile=4.3921933579313593e-07,
        upper whisker =2.9629147533720525e-06,
        lower whisker =9.726844078174102e-11
    },
    ] coordinates {};
    \addplot+[
    boxplot prepared={
        median        =0.06320832857748189,
        upper quartile=0.14757535247326686,
        lower quartile=0.05572403573492954,
        upper whisker =0.32093538573819097,
        lower whisker =0.046410950951204594
    },
    ] coordinates {};
    \addplot+[
    boxplot prepared={
        median        =0.05205291393250784,
        upper quartile=0.07086376955311702,
        lower quartile=0.04101778180199453,
        upper whisker =0.11981857258976411,
        lower whisker =0.033710648927289
    },
    ] coordinates {};
  \end{axis}
\end{tikzpicture}
\caption{The error in satisfying the bipendulum \glspl{ode} in Equation \eqref{eq:bipendel_system_equations} for a \gls{gp} (top) and \gls{lodegp} (bottom), for noisy training data. Smaller is better. The green dashed line shows the error of a solution in symbolic form. The low error shows that the \gls{lodegp} strictly satisfies the \glspl{ode}, up to numerical precision.}
\label{fig:bipendulum_ode_boxplot}
\end{figure}

Figure~\ref{fig:bipendulum_ode_boxplot} compares the error in satisfying the \gls{ode} of a \gls{lodegp} and a \gls{gp}.
The \gls{lodegp} is producing samples that satisfy the \gls{ode} with a median error\footnote{%
    The \glspl{lodegp} trains into a local minimum at its value $10^{-10}$, with high lengthscale and miniscule signal variance.
    This models constant zero behaviour which satisfies the \glspl{ode}, but does not fit the training data.
} of $\expnumber{2}{-6}$, for both differential equations.
In comparison, the \emph{symbolically computed solution} used to generate the training data, has an error of $\expnumber{1.75}{-7}$, only one order of magnitude better than the \gls{lodegp}.
Hence, we conclude that, up to numerical precision, the \gls{lodegp} produces samples that strictly satisfy the \gls{ode}, whereas regression models like \glspl{gp} do not satisfy the differential equations.
The \gls{lodegp} produces roughly $30000$ times more precise samples ($0.06$ to $\expnumber{2}{-6}$) and deviates only slightly from this error.

\begin{figure}
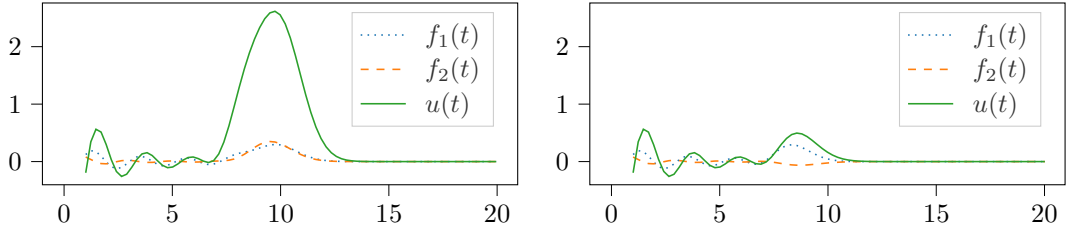

    \import{}{bipendulum_adam_mean.tex} 
    \import{}{bipendulum_classic_mean.tex}
    \caption{The posterior mean functions of a \gls{lodegp} (left) and a \gls{gp} (right) trained on the bipendulum data as described in the text.}
    \label{fig:bipendulum_adam_mean_comparison}
\end{figure}

Further we investigate the \gls{rmse} of the models by comparing their posterior mean to noiseless data in the training and evaluation intervals.
The \gls{lodegp} \gls{rmse} mostly performs similar compared to the \gls{gp} \gls{rmse} (see Table~\ref{tab:all_eval_results}).
For the training interval, the \gls{lodegp} achieves basically the same performance as the \gls{gp}. 
For the evaluation interval the performance worsens due to a \gls{se} \gls{gp} extrapolating to its (constant zero) prior mean.
The \gls{gp} can easily reach this mean zero, by smoothly modifying all channels without regard for the system equations.
Since the \gls{ode} solution also approaches zero, this leads to a small error in extrapolation.
The \gls{lodegp} has to satisfy the system equations to move the mean function back to mean zero.
Hence, the system produces a large spike\footnote{The third data channel is particularly large due to its additional factor of $g=9.81$, which has a different scale by an order of magnitude as the otherwise small data.} in the input channel $u(t)$, before the system can reach mean zero, as seen in Figure~\ref{fig:bipendulum_adam_mean_comparison}.

We also analyse the training runtime and see that the \gls{lodegp} is as fast as the \gls{gp} during the $\mathcal{O}(n^3)$ model inference and slower by a factor smaller than two in the $\mathcal{O}(n^2)$ covariance matrix calculation.

\subsection{Heating system - Controllable with parameters}\label{subsection:heating}

In this experiment we use a parameterized heating system to show that a \gls{lodegp} can learn physically interpretable system parameters during \gls{gp} hyperparameter training.
The controllable heating system is depicted in Figure~\ref{fig:airflow_system_visualization}.
It uses an input $u(t)$ to control a heating element $f_1(t)$ which exchanges heat with an object $f_2(t)$.
Two physical parameters $a$ and $b$ determine how quickly the heat moves between the objects according to the \glspl{ode}:
\begin{align*}
    f_1'(t) &= - a\cdot(f_1(t) - f_2(t)) + u(t)\\
    f_2'(t) &= - b\cdot(f_2(t) - f_1(t))
\end{align*}
We generate training data using a solution of the differential equation (see Figure~\ref{fig:airflow_system_visualization}) with parameter values $a=3, b=1$.
The \gls{lodegp} can be constructed with parameters in the \glspl{ode}.

\newcommand*\densitya{20}
\newcommand*\densityb{90}
\newcommand*\densityc{100}
\newcommand*\step{2}

\begin{figure}
    \center
    \begin{tikzpicture}[scale=0.5]
        \node[draw,cylinder,
        minimum width=0.5*4.25cm,
        minimum height=0.5*5.65cm,
        shape border rotate=90,
        aspect=1,
        bottom color=cyan!\densityb!red, 
        top color=cyan!\densitya!white!\densityc!red,
        anchor=after bottom] 
        at (-0.1,-0.1){};
        
        \draw[decoration={aspect=0.7, segment length=4mm, amplitude=2mm,coil},decorate,line width=3] (0.5,1) -- (3.75,1);
        \draw[line width=3] (0.5,1.05) -- (0.5,-1) -- (1.625,-1);
        \draw[line width=3] (3.7,1.05) -- (3.7,-1) -- (2.625,-1);
        \draw[line width=2] (1.875,-1) sin (2.0,-0.875) cos (2.125,-1) sin (2.25,-1.125) cos (2.375,-1);
        \draw[line width=2] (2.125,-1) circle (0.5);
        {
                \node at (2.75,-1.75) {$u(t)$};
                \node at (2.125,1.9) {$f_1(t)$};
                \node at (2.125,3.5) {$f_2(t)$};
        }
   \end{tikzpicture}
   \import{}{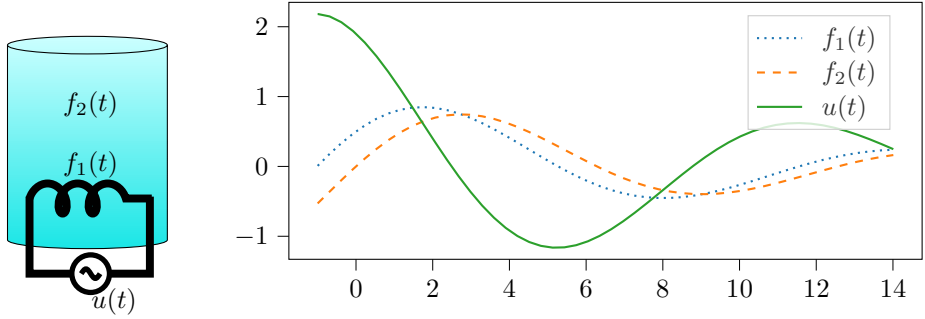}
    \caption{(left) A visualization of the heating system with input $u(t)$ and states $f_1(t), f_2(t)$. (right) A solution of the differential equations, which are used to create the data.}
    \label{fig:airflow_system_visualization}
\end{figure}

Learning the physical parameters $a$ and $b$ reconstructs the original values successfully with a maximal relative error of less than 2.8\% in ten training runs on data without noise.
After adding white noise to the data with standard deviation of 1\% of the maximal signal, the parameters were successfully reconstructed with a maximal relative error of 5.3\% in ten training runs.
Again, the \gls{lodegp} satisfies the original \glspl{ode} (with parameters $a=3$ and $b=1$) with an median error of $\expnumber{1}{-2}$, trained on the noisy data, we refer the reader to Appendix \ref{app:heating_details} for a visualization similar to Figure \ref{fig:bipendulum_posterior_our}.
This is bigger than in the previous example, as the \gls{lodegp} only uses approximate parameter values for $a$ and $b$, but still smaller than the error of a \gls{gp} of $0.20$.
Our interpretation of these results is that the \gls{lodegp} can learn the physically interpretable parameters.

\subsection{Three tank system - Non-controllable}
\begin{figure}
    \begin{tikzpicture}[scale=0.65]
  \def\tankWidth{2}
  \def\tankHeight{2}
  \def\leftTankPipeDistance{1.5*\tankHeight}
  \def\leftMinPipeDistance{\leftTankPipeDistance-0.5}
  \def\rightTankPipeDistance{1.2+\leftTankPipeDistance}
  \def\rightMinPipeDistance{\rightTankPipeDistance-0.5}
  \def\pipeWidth{0.2*\tankWidth}
  \coordinate (left_tank_x_center) at (0,0);
  \coordinate (middle_tank_x_center) at ($(left_tank_x_center)+(1.2*\tankWidth,0)$);
  \coordinate (right_tank_x_center) at ($(middle_tank_x_center)+(1.2*\tankWidth,0)$);

  \draw ($(left_tank_x_center)+(-0.5*\tankWidth,0)$) arc [
        start angle=0,
        end angle=180,
        x radius=-0.5*\tankWidth,
        y radius=-0.1*\tankHeight
    ]-- ($(left_tank_x_center)+(0.5*\tankWidth,\tankHeight)$)
    arc [
        start angle=0,
        end angle=540,
        x radius=0.5*\tankWidth,
        y radius=-0.1*\tankHeight
    ] --($(left_tank_x_center)+(-0.5*\tankWidth,0)$) ;

  \draw ($(middle_tank_x_center)+(-0.5*\tankWidth,0)$) arc [
        start angle=0,
        end angle=180,
        x radius=-0.5*\tankWidth,
        y radius=-0.1*\tankHeight
    ]-- ($(middle_tank_x_center)+(0.5*\tankWidth,\tankHeight)$)
    arc [
        start angle=0,
        end angle=540,
        x radius=0.5*\tankWidth,
        y radius=-0.1*\tankHeight
    ] --($(middle_tank_x_center)+(-0.5*\tankWidth,0)$);

  \draw ($(right_tank_x_center)+(-0.5*\tankWidth,0)$) arc [
        start angle=0,
        end angle=180,
        x radius=-0.5*\tankWidth,
        y radius=-0.1*\tankHeight
    ]-- ($(right_tank_x_center)+(0.5*\tankWidth,\tankHeight)$)
    arc [
        start angle=0,
        end angle=540,
        x radius=0.5*\tankWidth,
        y radius=-0.1*\tankHeight
    ] --($(right_tank_x_center)+(-0.5*\tankWidth,0)$);

  \draw ($(left_tank_x_center)+(-0.7*\tankWidth, \leftTankPipeDistance)$)
      -- ($(left_tank_x_center)+(-0.5*\pipeWidth, \leftTankPipeDistance)$)
      -- ($(left_tank_x_center)+(-0.5*\pipeWidth, \leftMinPipeDistance)$)
    arc [
        start angle=0,
        end angle=180,
        x radius=-0.5*\pipeWidth,
        y radius=-0.2*\pipeWidth
    ] -- ($(left_tank_x_center)+(0.5*\pipeWidth, \leftTankPipeDistance)$)
      -- ($(middle_tank_x_center)+(-0.5*\pipeWidth, \leftTankPipeDistance)$)
      -- ($(middle_tank_x_center)+(-0.5*\pipeWidth, \leftMinPipeDistance)$)
    arc [
        start angle=0,
        end angle=180,
        x radius=-0.5*\pipeWidth,
        y radius=-0.2*\pipeWidth
    ] -- ($(middle_tank_x_center)+(0.5*\pipeWidth, \pipeWidth+\leftTankPipeDistance)$)
      -- ($(left_tank_x_center)+(-0.7*\tankWidth, \pipeWidth+\leftTankPipeDistance)$);

  \draw ($(right_tank_x_center)+(0.7*\tankWidth, \rightTankPipeDistance)$)
      -- ($(right_tank_x_center)+(0.5*\pipeWidth, \rightTankPipeDistance)$)
      -- ($(right_tank_x_center)+(0.5*\pipeWidth, \rightMinPipeDistance)$)
    arc [
        start angle=0,
        end angle=180,
        x radius=0.5*\pipeWidth,
        y radius=-0.2*\pipeWidth
    ] -- ($(right_tank_x_center)+(-0.5*\pipeWidth, \rightTankPipeDistance)$)
      -- ($(middle_tank_x_center)+(0.5*\pipeWidth, \rightTankPipeDistance)$)
      -- ($(middle_tank_x_center)+(0.5*\pipeWidth, \rightMinPipeDistance)$)
    arc [
        start angle=0,
        end angle=180,
        x radius=0.5*\pipeWidth,
        y radius=-0.2*\pipeWidth
    ] -- ($(middle_tank_x_center)+(-0.5*\pipeWidth, \pipeWidth+\rightTankPipeDistance)$)
      -- ($(right_tank_x_center)+(0.7*\tankWidth, \pipeWidth+\rightTankPipeDistance)$);
  \node (left_tank_text) at ($(left_tank_x_center)+(0, 0.1)$) {$f_1(t)$};
  \node (middle_tank_text) at ($(middle_tank_x_center)+(0, 0.1)$) {$f_2(t)$};
  \node (right_tank_text) at ($(right_tank_x_center)+(0, 0.1)$) {$f_3(t)$};
  \node (left_pipe_text) at ($(left_tank_x_center)+(-0.9*\tankWidth, \leftTankPipeDistance + 0.5*\pipeWidth)$) {$u_1(t)$};
  \node (right_pipe_text) at ($(right_tank_x_center)+(0.9*\tankWidth, \rightTankPipeDistance + 0.5*\pipeWidth)$) {$u_2(t)$};
    \end{tikzpicture}
    \import{}{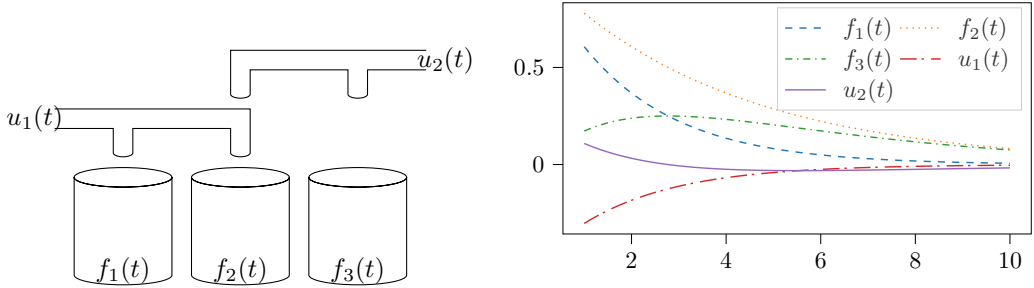}
    \caption{(left) A sketch of the three tank system 
    and (right) a solution of the system.}
    \label{fig:three_tank_sketch}
\end{figure}

We conclude the experiments with a non-controllable fluid system where the water level in three tanks is changed by two pipes.
The system is non-controllable due to pipes' overlap over the center tank, whose changes directly affect the other tanks.
This system requires multiple non-zero covariance functions in the latent \gls{gp} to describe both the non-controllable subsystem and the two degrees of freedom, which corresponds to the columns of the \glspl{snf} matrix $D$ via Lemma \ref{lemma:base_cov_table}:
\begin{align*}
    D = \begin{bmatrix}
        1&  0&  0&  0&  0\\
        0&  1&  0&  0&  0\\
        0&  0& -\partial_t&  0&  0
    \end{bmatrix}
\end{align*}

We use a solution to the system of \glspl{ode} (see Figure~\ref{fig:three_tank_sketch}) to generate 25 datapoints, to which we add white noise with standard deviation of 10\% of the maximal signal.

The system \glspl{ode} are again strictly satisfied by the \gls{lodegp} with a median error of $\expnumber{1}{-5}$ in each \gls{ode}.
We refer the reader to Appendix \ref{app:three_tank_details} for a visualization similar to Figure \ref{fig:bipendulum_posterior_our}.
Hence, \gls{lodegp} can handle larger, non-controllable systems of \glspl{ode}, despite high noise.

\section{Conclusion}
In this paper we have introduced an algorithmical approach to automatically and symbolically create \gls{lodegp}, a class of covariance functions for \glspl{gp} such that their realizations strictly follow a given system of linear homogeneous \glspl{ode}.
We have proven that this approach is mathematically sound and verified its effectiveness in experiments.
We have demonstrated that the learned posterior mean strictly satisfies the given system of \glspl{ode} up to numerical precision and discussed the consequences of this strict behavior e.g.\ in their extrapolation, leading to drastic behavior when both following the \glspl{ode} and going back to the prior mean. 
Additionally, we automatically trained the physically interpretable parameters of the system and were able to reconstruct their original values with low relative error.
The runtime of the \gls{lodegp} is still asymptotically dominated by the $\mathcal{O}(n^3)$ Cholesky decomposition of the covariance matrix, despite bigger constants in the $\mathcal{O}(n^2)$ part of constructing the covariance matrix and a (usually very small) constant time precomputation of the covariance function.
Of course, approximate \glspl{gp} are applicable to our covariance functions.
Beyond applications where the dynamic behavior is known through corresponding \glspl{ode}, the combination with kernel search methods \cite{kim2018scaling, duvenaud2013structure, huwel2021automated, berns20213cs, berns2020automatic} has the potential to detect unknown dynamical behavior in data by systematic exploration and use these findings to generate new interpretable knowledge about the data through the corresponding system.

\begin{ack}
    This research was supported by the research training group ``Dataninja'' (Trustworthy AI for Seamless Problem Solving: Next Generation Intelligence Joins Robust Data Analysis) funded by the German federal state of North Rhine-Westphalia.

    We want to thank the anonymous reviewers for their valuable feedback.
\end{ack}

\bibliographystyle{plainnat}
\bibliography{bibliography}

\newpage
\appendix
\section*{Appendices}
\renewcommand{\thesubsection}{\Alph{subsection}}

\subsection{Constructing and training a LODE-GP}
The construction of a \gls{lodegp}, as discussed in the proof of Theorem~\ref{theorem:gp_model_from_snf}, involves the following steps:
\begin{enumerate}
    \item Calculate the \gls{snf} symbolically using SageMath \cite{sagemath}
    \item Construct the covariance function for the \gls{gp}-prior $h \sim \GP(\mathbf{0}, k)$, according to Lemma~\ref{lemma:base_cov_table} and the last line of Equation~\ref{eq:DiffEq_Lf_eq_0} 
    \item Calculate the final \gls{lodegp} through the pushforward with the matrix $V$, including the matrix multiplication $V\cdot k\cdot V'$ and applying the derivatives to $k$
\end{enumerate}
This construction gives us our \gls{lodegp}, which we train using the standard \gls{gp} training procedure.
All \gls{se} kernels that are part of the \glspl{lodegp} have randomly initialized signal variance $\sigma$ and lengthscale $\ell$ values chosen uniformly from $[-3, 3]$, which are passed through an $\exp$, during calculation, to ensure numerical stability (e.g.\ preventing negative lengthscales) and smooth training.
We use an Adam optimizer with a learning rate of $0.1$ for 300 training iterations in all experiments, further we use the GPyTorch multitask likelihood, since the \gls{lodegp} inherits the class of a GPyTorch multitask \gls{gp}.
We reduced the noise constraint for the likelihood from $\expnumber{1}{-4}$ to $\expnumber{1}{-10}$.
GPyTorch uses the softplus operation to ensure the constraint.
We performed calculations with 64 bit precision.

Training and evaluation was done on our server equipped with a Intel(R) Core(TM) i9-10900K CPU @ 3.70GHz and 64GB RAM running at 3200MHz. 
All calculations were done on the CPU.

Training was done on 25 datapoints and was repeated ten times for noisy and noise-free, for both \gls{lodegp} and \gls{gp}.

At no point in training or evaluation is any data or \gls{gp} output scaled, i.e.\ we use all the data as is.

\subsection{Additional details on experiments and results}\label{app:experiment_details}

We compare the values for the \gls{gp} and \gls{lodegp} parameters across the three experiments in Table~\ref{tab:parameters}.
The values for the parameters are their actual values as used in the calculation, i.e.\ after applying the softplus and exp functions.
The behaviour discussed in Section~\ref{sec:evaluation} is clearly visible in the maximum \gls{lodegp} lengthscale for the bipendulum, where it sometimes learns a basically constant function.
For the bipendulum and the three tank experiments, the \gls{lodegp} generally learns a higher lengthscale, which translates to a smoother behaviour in the outputs.
\begin{table}
    \caption{The trained hyperparameters for the bipendulum (top), heating (middle) and three tank (bottom) experiments, for the noisy and noise-free settings. The \gls{gp} trains three, resp.\ five, signal variances, for simpler presentation we show the mean of the signal variance parameters. For the \gls{lodegp} three tank system we also use the mean of the two \gls{se} kernels that are created.}
\begin{tabular}{lllllll}
	\toprule
    & Parameter & Median                   & Std. deviation & Minimum & Maximum\\
\midrule
    \gls{lodegp} noisy& noise                    &  6.7521e-05  & 0.000149199 &  4.91994e-05 &  0.000470599 \\
& signal variance                                &  0.0617262   & 0.0351961   &  7.22543e-06 &  0.0879048   \\
& lengthscale                                    &  1.24498     & 3.62295     &  1.18036     & 13.4538      \\
 \midrule
    \gls{lodegp} no noise& noise                 &  0.000306269 & 0.000178327 &  1.7952e-11  &  0.000413706 \\
& signal variance                                &  0.000298894 & 0.109973    &  6.91032e-06 &  0.354441    \\
& lengthscale                                    &  1.82798     & 0.86266     &  0.69606     &  3.54208     \\
 \midrule
    \gls{gp} noisy& noise                        &  8.60043e-05 & 1.53343e-05 &  4.62593e-05 &  9.75798e-05 \\
& signal variance                                &  0.00222937  & 0.00135264  &  0.000539298 &  0.0056802   \\
& lengthscale                                    &  0.786804    & 0.0557031   &  0.709304    &  0.86083     \\
 \midrule
    \gls{gp} no noise& noise                     &  2.66449e-09 & 2.13689e-08 &  5.74964e-11 &  5.48067e-08 \\
& signal variance                                &  0.0436976   & 0.0591215   &  3.70462e-05 &  0.152265    \\
& lengthscale                                    &  1.25061     & 0.0776317   &  1.09688     &  1.34608     \\
\midrule
\midrule
    \gls{lodegp} noisy& noise                    &  0.000186034 & 3.64772e-05 &  0.000135545 &  0.000243457 \\
& signal variance                                & 10.0713      & 4.51968     &  0.412769    & 15.3753      \\
& lengthscale                                    &  5.46155     & 0.553268    &  3.7435      &  5.86303     \\
\midrule
    \gls{lodegp} no noise& noise                 &  1.41396e-06 & 1.97306e-06 &  4.83248e-10 &  5.48139e-06 \\
& signal variance                                &  6.43588     & 0.520707    &  5.76422     &  7.13915     \\
& lengthscale                                    &  5.06497     & 0.109923    &  4.87297     &  5.2045      \\
 \midrule
    \gls{gp} noisy& noise                        &  0.000169646 & 2.93415e-05 &  0.000132707 &  0.000218817 \\
& signal variance                                &  1.26965     & 0.763973    &  0.542204    &  3.03235     \\
& lengthscale                                    &  4.20254     & 0.198991    &  3.8563      &  4.47665     \\
 \midrule
    \gls{gp} no noise& noise                     &  1.61976e-11 & 5.0974e-13  &  1.57147e-11 &  1.72362e-11 \\
& signal variance                                &  6.29899     & 1.27418     &  3.36005     &  7.5202      \\
& lengthscale                                    &  5.08887     & 0.175657    &  4.77775     &  5.26459     \\
\midrule
\midrule
    \gls{lodegp} noisy& noise                    &  0.00287473  &  0.000523729 &  0.00238355  &   0.00393084  \\
& signal variance                                &  0.58266     & 0.192348    &  0.323803    &  0.913425    \\
& lengthscale                                    &  5.41985     & 1.00242     &  4.31772     &  7.55746     \\
 \midrule
    \gls{lodegp} no noise& noise                 &  1.18695e-11 &  6.09864e-13 &  1.05086e-11 &   1.23305e-11 \\
& signal variance                                & 44.2332      & 5.66572     & 32.4225      & 51.8143      \\
& lengthscale                                    &  6.66897     & 0.116321    &  6.42639     &  6.8065      \\
 \midrule
    \gls{gp} noisy& noise                        &  0.00281939  &  0.000507194 &  0.00176329  &   0.00377263  \\
& signal variance                                &  0.00104692  &  0.000369078 &  0.000433981 &   0.00150992  \\
& lengthscale                                    &  5.52465     &  1.15024     &  3.01939     &   6.5203      \\
 \midrule
    \gls{gp} no noise& noise                     &  1.6506e-11  &  3.734e-13   &  1.58496e-11 &   1.72428e-11 \\
& signal variance                                &  0.00304028  &  0.00140299  &  0.00170012  &   0.006194    \\
& lengthscale                                    &  4.42061     &  0.113491    &  4.20928     &   4.69174     \\
\bottomrule
\end{tabular}
\label{tab:parameters}
\end{table}

\begin{table}[h]
    \centering
    \caption{Median number of eigenvalues greater than zero (i.e.\ eigenvalue $\lambda > \expnumber{1}{-6}$) for the experiments with the \gls{lodegp} and the \gls{gp}. The total number of eigenvalues for the bipendulum/heating and three tank are 3000 and 5000, respectively.}
    \begin{tabular}{llc}
        \toprule
        && eigenvalues greater zero\\
        \midrule
        Bipendulum & \gls{lodegp} & 29\\
                   & \gls{gp} & 76\\
        Heating & \gls{lodegp} & 14\\
                & \gls{gp} & 39\\
        Three tank & \gls{lodegp} & 18.5\\
                   & \gls{gp} & 32.5\\
                   \bottomrule
    \end{tabular}
    \label{tab:eigenvalues_greater_zero}
\end{table}
Analyzing the eigenvalues for the \gls{lodegp} and \gls{gp} shows that the \gls{lodegp} produces roughly half as many non-zero eigenvalues (i.e.\ eigenvalue $\lambda > \expnumber{1}{-6}$) compared to the \gls{gp}, as shown in Table~\ref{tab:eigenvalues_greater_zero}.
This behaviour is expected since the \gls{lodegp} is underlying additional strong constraints due to the inductive bias, which is reflected in its eigenvalues.

\subsubsection{Calculating the ODE error}

To calculate the error in satisfying the \glspl{ode} via finite differences, we generate uniformly distributed datapoints in the evaluation interval and add duplicate points, shifted by a step of $\Delta = 1e-3$.
We generate either $500$ or $333$ datapoints, dependent on whether only the first or also the second derivative is necessary, ensuring that always a total of $1000$ datapoints are created.

We pass the data to the \gls{gp} resp.\ \gls{lodegp} and extract the resulting mean function.
The first and second derivative are then approximated using forward difference.

To ensure numerical stability in the calculation of the second derivative, we pass the first order forward difference result through a low-pass filter, removing high frequency noise from the result.

To finally calculate the \gls{ode} error, we insert the finite difference values for the corresponding derivatives in the original \glspl{ode} and average over the error for each \gls{ode}.

\subsubsection{Bipendulum}
We illustrate how a \gls{lodegp} covariance function might look like by showing the (1,1) entry for the bipendulum \gls{lodegp} covariance function:
\begin{align*}
\sigma^2\exp\left(-\frac{(x_1-x_2)^2}{2\ell^2}\right)\cdot
\left(
\frac{(x_1-x_2)^4}{\ell{^8}} - \frac{6\cdot(x_1-x_2)^2}{\ell{^6}} + \frac{3 + g\cdot(x_1-x_2)^2}{\ell{^4}} - \frac{g}{\ell{^2}} + \frac{g^2}{4}
\right)
\end{align*}

\paragraph{System details}
Since the bipendulum has been thoroughly discussed in the paper, we just briefly mention the system equations, operator matrix and the \gls{snf}:
\begin{align*}
   \begin{bmatrix}
    x''(t) + \ell_1f_1''(t) + gf_1(t) \\
    x''(t) + \ell_2f_2''(t) + gf_2(t) 
\end{bmatrix} = \mathbf{0} = 
\underbrace{
    \begin{bmatrix}
        \partial_t^2 + \frac{g}{\ell_1} & 0 & -\frac{1}{\ell_1} \\
        0 & \partial_t^2 + \frac{g}{\ell_1} & -\frac{1}{\ell_2}
    \end{bmatrix}}_A
    \cdot
    \begin{bmatrix}
        f_1(t) \\
        f_2(t) \\
        u(t)
    \end{bmatrix}
\end{align*}
\begin{align*}
    &\text{case } \ell_1=1 \neq \ell_2=2\text{:}\\
    &\underbrace{
    \begin{bmatrix}
    1 & 0 \\
    -\frac{1}{2} & 1
    \end{bmatrix}
    }_U
    \underbrace{
    \begin{bmatrix}
        \partial_t^2 + g & 0 & -1 \\
        0 & \partial_t^2 + \frac{g}{2} & -\frac{1}{2}
    \end{bmatrix}
    }_A
    \underbrace{
    \begin{bmatrix}
            0 & -\frac{4}{g} & \frac{2\partial_t^{2}+g}{2} \\
            0 & -\frac{2}{g} & \frac{\partial_t^{2}+g}{2} \\
        -1 & -\frac{4\partial_t^{2}+4g}{g}  & (\partial_t^2+\frac{g}{2})(\partial_t^2+g)
    \end{bmatrix}
    }_V
    =
    \underbrace{
    \begin{bmatrix}
    1 & 0 & 0 \\
    0 & 1 & 0
    \end{bmatrix}
    }_D
\end{align*}

To generate the data we use the following solution to the \gls{ode} (cf.\ Figure~\ref{fig:bipendulum_posterior_our}):
\begin{align}\label{eq:bipendulum_ode_solution}
    \begin{bmatrix}
        f_1(t)\\
        f_2(t)\\
        u(t)
    \end{bmatrix} = 
    \begin{bmatrix}
    -\frac{41 \, \sin\left(3 \, t\right)}{100 \, {\left(t + 1\right)}} - \frac{3 \, \cos\left(3 \, t\right)}{5 \, {\left(t + 1\right)}^{2}} + \frac{\sin\left(3 \, t\right)}{5 \, {\left(t + 1\right)}^{3}}\\
\frac{81 \, \sin\left(3 \, t\right)}{2000 \, {\left(t + 1\right)}} - \frac{3 \, \cos\left(3 \, t\right)}{10 \, {\left(t + 1\right)}^{2}} + \frac{\sin\left(3 \, t\right)}{10 \, {\left(t + 1\right)}^{3}}\\
-\frac{3321 \, \sin\left(3 \, t\right)}{10000 \, {\left(t + 1\right)}} + \frac{987 \, \cos\left(3 \, t\right)}{500 \, {\left(t + 1\right)}^{2}} - \frac{3929 \, \sin\left(3 \, t\right)}{500 \, {\left(t + 1\right)}^{3}} - \frac{36 \, \cos\left(3 \, t\right)}{5 \, {\left(t + 1\right)}^{4}} + \frac{12 \, \sin\left(3 \, t\right)}{5 \, {\left(t + 1\right)}^{5}}
    \end{bmatrix}
\end{align}

\begin{figure}
    \import{}{bipendulum_posterior_our.tex}
    \import{}{BipendulumSimplified_Our_NoNoise_Adam_300_0.001_1.0-6.0-25_1.0-11.0-1000.tex}
    \import{}{bipendulum_sample.tex}
    \caption{The posterior \gls{lodegp} models for the bipendulum system, trained on noisy data (top) and noise-free data (bottom). The black stars indicate the datapoints (with or without noise), the red line is the solution to the \glspl{ode}, the blue dashed line is the \glspl{lodegp} posterior mean, the transparent blue area is the $2\sigma$ confidence interval. In the bottom plot, a sample from the GP is shown.}
    \label{fig:bipendulum_posterior_our}
\end{figure}

\paragraph{Training details}
To train a \gls{lodegp} we generate uniformly spaced training data from Equation~\ref{eq:bipendulum_ode_solution} in the interval $[1, 6]$ to which we add noise of the form $0.012 \cdot \sigma_n$ for $\sigma_n \sim \mathcal{N}(0, 1)$, i.e.\ noise with standard deviation $0.012$ ($2$\% of the maximal signal value). 
The data is shown in Figure~\ref{fig:bipendulum_posterior_our}.
To validate our results, e.g.\ using \gls{rmse}, we generate uniformly spaced evaluation data from Equation~\ref{eq:bipendulum_ode_solution} in the interval $[1, 11]$, without adding noise. 

\subsubsection{Heating}\label{app:heating_details}
\paragraph{System details}

Original system equations:
\begin{align*}
    \begin{bmatrix}
    - f_1'(t) - a\cdot(f_1(t) - f_2(t)) + u(t)\\
    - f_2'(t) - b\cdot(f_2(t) - f_1(t))
\end{bmatrix}
 = \mathbf{0} 
 = \underbrace{\begin{bmatrix}
        \partial_t + a & -a & -1\\
        -b & \partial_t + b & 0
\end{bmatrix}}_A
    \cdot
    \begin{bmatrix}
        f_1(t)\\
        f_2(t)\\
        u(t)
    \end{bmatrix}
\end{align*}

The \gls{snf} of the \glspl{ode} with all relevant matrices:
\begin{align*}
\underbrace{
\begin{bmatrix}
1 & 0 & 0 \\
0 & 1 & 0
\end{bmatrix}}_D
=
\underbrace{
\begin{bmatrix}
0 & \frac{-1}{b} \\
b & \partial_t + a
\end{bmatrix}}_U
\underbrace{
    \begin{bmatrix}
        \partial_t + a & -a & -1\\
        -b & \partial_t + b & 0
    \end{bmatrix}
}_A
\underbrace{
\begin{bmatrix}
1 & 0 & \partial_t + b \\
0 & 0 & b \\
0 & \frac{-1}{b} & \partial_t^{2} + \left(b + a\right)\partial_t  
\end{bmatrix}}_V
\end{align*}

\begin{figure}
    \import{}{heating_our_noisy_posterior.tex}
    \import{}{heating_our_noisefree_posterior.tex}
    \import{}{heating_sample.tex}
    \caption{The posterior \gls{lodegp} models for the heating system, trained on noisy data (top) and noise-free data (bottom). The black stars indicate the datapoints (with or without noise), the red line is the solution to the \glspl{ode}, the blue dashed line is the \glspl{lodegp} posterior mean, the transparent blue area is the $2\sigma$ confidence interval. In the bottom plot, a sample from the GP is shown.}
    \label{fig:heating_posterior_our}
\end{figure}

To generate data, we use the following solution to the heating \glspl{ode} (cf.\ Figure~\ref{fig:heating_posterior_our}).
\begin{align}\label{eq:heating_ode_solution}
    \begin{bmatrix}
        f_1(t)\\
        f_2(t)\\
        u(t)
    \end{bmatrix}
    =
    \begin{bmatrix}
        \frac{1}{2} \, \cos\left(\frac{1}{2} \, t\right) \exp\left(-\frac{1}{10} \, t\right) + \frac{9}{10} \, \exp\left(-\frac{1}{10} \, t\right) \sin\left(\frac{1}{2} \, t\right)\\
    \exp\left(-\frac{1}{10} \, t\right) \sin\left(\frac{1}{2} \, t\right)\\
    \frac{19}{10} \, \cos\left(\frac{1}{2} \, t\right) \exp\left(-\frac{1}{10} \, t\right) - \frac{16}{25} \, \exp\left(-\frac{1}{10} \, t\right) \sin\left(\frac{1}{2} \, t\right)
\end{bmatrix}
\end{align}

\paragraph{Training details}
To train a \gls{lodegp} we generate uniformly spaced training data from Equation~\ref{eq:heating_ode_solution} in the interval $[-5, 5]$ to which we add noise of the form $0.02 \cdot \sigma_n$ for $\sigma_n \sim \mathcal{N}(0, 1)$, i.e.\ noise with standard deviation $0.02$ ($1$\% of the maximal signal value). 
The data is shown in Figure~\ref{fig:heating_posterior_our}.
To validate our results, e.g.\ using \gls{rmse}, we generate uniformly spaced evaluation data from Equation~\ref{eq:heating_ode_solution} in the interval $[-9, 9]$, without adding noise. 

\paragraph{Additional results}
We present the exact parameter values trained by the \gls{lodegp} in Table~\ref{tab:heating_parameters_relative_error}.
It can be seen that the \gls{lodegp} consistently learn the system parameters $a$ and $b$ with a small relative error, even despite noise.
\begin{table}
	\caption{The trained parameter values for $a$ and $b$ by the \gls{lodegp} and their relative error from their actual value, for training data with noise (top) and without noise (bottom).  Smaller rel.\ errors are better.}
    \centering
\begin{tabular}{llll}
    \toprule
    $a$ & $b$ & relative error $a$ & relative error $b$\\
    \midrule
 2.92218 & 0.983648 & 0.0259396     & 0.0163522    \\
 2.9831  & 0.997748 & 0.00563382    & 0.00225164   \\
 2.99933 & 0.991803 & 0.00022441    & 0.00819681   \\
 3.00912 & 1.01786  & 0.00304097    & 0.0178581    \\
 2.92706 & 0.984112 & 0.0243126     & 0.0158881    \\
 2.92016 & 0.968457 & 0.0266129     & 0.0315435    \\
 2.96962 & 0.989062 & 0.0101259     & 0.0109384    \\
 3.00482 & 0.990055 & 0.00160732    & 0.00994511   \\
 3.00571 & 1.00561  & 0.00190434    & 0.00560704   \\
 3.02601 & 0.998836 & 0.0086684     & 0.00116403   \\
 \midrule
 3.00065 & 1.00025  & 0.000217934   & 0.000254375  \\
 3.02079 & 0.989905 & 0.00693057    & 0.0100952    \\
 3.00042 & 1.00132  & 0.000139129   & 0.00132404   \\
 3.00002 & 1.00177  & 0.00000558486 & 0.00176581   \\
 2.97293 & 1.03124  & 0.00902484    & 0.0312367    \\
 2.98821 & 1.01434  & 0.00392941    & 0.0143379    \\
 3.0007  & 0.99991  & 0.00023333    & 0.0000902065 \\
 2.98481 & 1.02509  & 0.00506387    & 0.0250855    \\
 2.99362 & 1.01282  & 0.00212785    & 0.0128176    \\
 3.00045 & 1.00002  & 0.000149713   & 0.0000185651 \\
 \bottomrule
\end{tabular}
 \label{tab:heating_parameters_relative_error}
 \end{table}

\newpage
\subsubsection{Three tank}\label{app:three_tank_details}

Original system equations:
\begin{align*}
    \begin{bmatrix}
        -f_1'(t) + u_1(t)\\
        -f_2'(t) + u_1(t) + u_2(t)\\
        -f_3'(t) + u_2(t)
\end{bmatrix}
 = \mathbf{0} 
 = \underbrace{\begin{bmatrix}
-\partial_t & 0 & 0 & 1 & 0 \\
0 & -\partial_t & 0 & 1 & 1 \\
0 & 0 & -\partial_t & 0 & 1
\end{bmatrix}}_A
    \cdot
    \begin{bmatrix}
        f_1(t)\\
        f_2(t)\\
        f_3(t)\\
        u_1(t)\\
        u_2(t)
    \end{bmatrix}
\end{align*}

The \gls{snf} of the \glspl{ode} with all relevant matrices:
\begin{align*}
\underbrace{\begin{bmatrix}
1 & 0 & 0 \\
-1 & 1 & 0 \\
1 & -1 & 1
\end{bmatrix}}_U
\underbrace{\begin{bmatrix}
-\partial_t & 0 & 0 & 1 & 0 \\
0 & -\partial_t & 0 & 1 & 1 \\
0 & 0 & -\partial_t & 0 & 1
\end{bmatrix}}_A
\underbrace{\begin{bmatrix}
0 & 0 & 0 & -1 & 0 \\
0 & 0 & 0 & 0 & -1 \\
0 & 0 & 1 & 1 & -1 \\
1 & 0 & 0 & -\partial_t & 0 \\
0 & 1 & 0 & \partial_t & -\partial_t
\end{bmatrix}}_V
=
\underbrace{\begin{bmatrix}
1 & 0 & 0 & 0 & 0 \\
0 & 1 & 0 & 0 & 0 \\
0 & 0 & -\partial_t & 0 & 0
\end{bmatrix}}_D
\end{align*}

\begin{landscape}
\begin{figure}
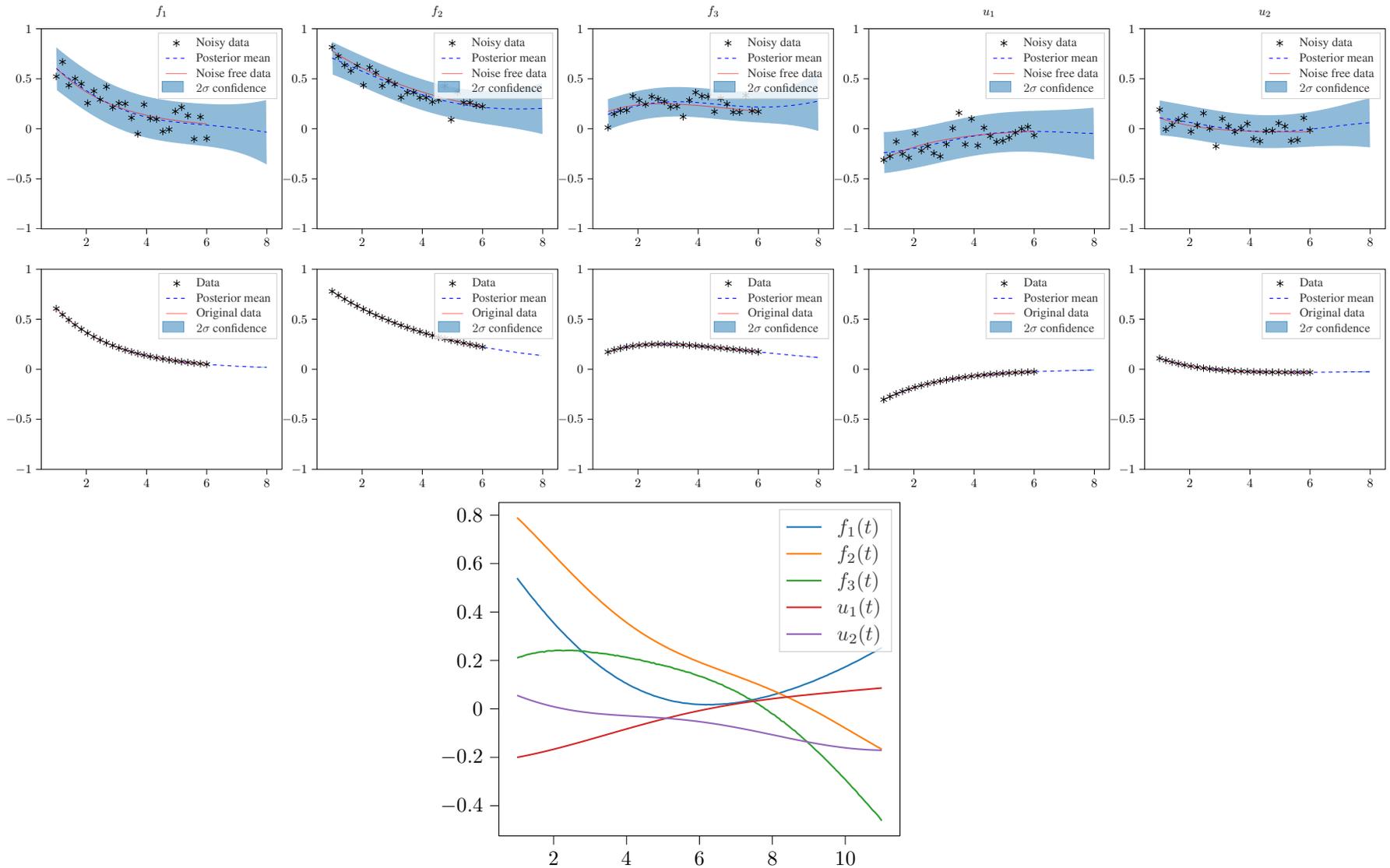

    \centering
    \import{}{ThreeTank_Our_GlobalNoise-0.08_TaskNoises-0.0-0.0-0.0-0.0-0.0_Adam_300_0.001_1.0-6.0-25_1.0-11.0-1000.tex}
    \import{}{ThreeTank_Our_GlobalNoise-0.0_TaskNoises-0.0-0.0-0.0-0.0-0.0_Adam_300_0.001_1.0-6.0-25_1.0-11.0-1000.tex}
    \import{}{threeTank_sample.tex}
    \caption{The posterior \gls{lodegp} models for the three tank system, trained on noisy data (top) and noise-free data (bottom). The black stars indicate the datapoints (with or without noise), the red line is the solution to the \glspl{ode}, the blue dashed line is the \glspl{lodegp} posterior mean, the transparent blue area is the $2\sigma$ confidence interval. In the bottom plot, a sample from the GP is shown.}
    \label{fig:three_tank_posterior_our}
\end{figure}
\end{landscape}

We use the following solution to the \glspl{ode} (cf.\ Figure~\ref{fig:three_tank_posterior_our}).
\begin{align}\label{eq:three_tank_solution}
    \begin{bmatrix}
        f_1(t)\\
        f_2(t)\\
        f_3(t)\\
        u_1(t)\\
        u_2(t)
    \end{bmatrix}
    =
    \begin{bmatrix}
        \exp (-\frac{t}{2})\\
        \exp (-\frac{t}{4})\\
        \exp (-\frac{t}{4})-\exp(-\frac{t}{2})\\
        -\frac{\exp (-\frac{t}{2})}{2}\\
        -\frac{\exp (-\frac{t}{4})}{4} + \frac{\exp (-\frac{t}{2})}{2}
    \end{bmatrix}
\end{align}

\paragraph{Training details}
To train a \gls{lodegp} we generate uniformly spaced training data from Equation~\ref{eq:three_tank_solution} in the interval $[1, 6]$ to which we add noise of the form $0.08 \cdot \sigma_n$ for $\sigma_n \sim \mathcal{N}(0, 1)$, i.e.\ noise with standard deviation $0.08$ ($10$\% of the maximal signal value). 
The data is shown in Figure~\ref{fig:three_tank_posterior_our}.
To validate our results, e.g.\ using \gls{rmse}, we generate uniformly spaced evaluation data from Equation~\ref{eq:three_tank_solution} in the interval $[1, 11]$, without adding noise. 

For the three tank system we hard coded the case $-x = x$, which is a legal operation since we can take out the $-1$ as part of a left matrix multiplication and integrate it in the $U$ matrix, which does not influence the \gls{lodegp}.

\paragraph{Comparison to Latent Force Model}
We compare our \gls{lodegp} with the \gls{lfm} introduced by \cite{alvarez2009latent}.
To do so we set the variables from Equation 3 in \cite{alvarez2009latent} as follows:
$D_q = 0$, $B_q = 0$, $S_rq=\begin{bmatrix} -1 & 0 \\ -1 & -1 \\ 0 & -1 \end{bmatrix}$.
Further we calculate the covariance function as the solution of the integral $k = \int_0^{t_2}\int_0^{t_1} \exp^{-(x_1 - x_2)^2} dx_1 dx_2$, effectively setting $\ell = 1$ and $\sigma = 1$.
Following the steps of \cite{alvarez2009latent}, we get a \gls{gp} that can estimate the solution of the three tank systems differential equations.

The resulting \gls{gp} marginalizes the (in their model considered) latent function $u_1$ and $u_2$ and only considered the three data channels $f_1$, $f_2$, and $f_3$. 
To calculate the \gls{ode} error, we inserted the original data for the channels $u_1$  and $u_2$ into the calculation.

For a fair comparison with our \gls{lodegp}, we also marginalize $u_1$ and $u_2$ there.
Similarly as for the \gls{lfm}, we have set $\ell = 1$ and $\sigma = 1$.

The resulting \gls{ode} errors of the two models are as shown in Table~\ref{tab:lfm_comparison_results}.
The performance of the marginalized \gls{lodegp} is better but comparable to the \gls{lfm}.
The performance of the full \gls{lodegp}, having learned all 5 channels and also set all lengthscales and signal variances to $1$, shows a signifcant increase in performance.

\begin{table}[h]
    \caption{The \gls{ode} error of the \gls{lfm}, the small \gls{lodegp}, and the full \gls{lodegp}. Smaller is better.}
        \centering
    \begin{tabular}{cccc}
		\toprule
        & \gls{ode} 1 error & \gls{ode} 2 error & \gls{ode} 3 error \\
        \midrule
        (marginalized) \gls{lfm} & 0.042816& 0.507017& 0.013084 \\
        \midrule
        marginalized \gls{lodegp} & 0.031331& 0.025065& 0.008595\\
        \midrule
        full \gls{lodegp} &1.210e-05& 1.285e-05& 1.081e-05\\
        \bottomrule
    \end{tabular}
\label{tab:lfm_comparison_results}
\end{table}

\subsection{Discussion of the code and runtime}
We use GPyTorch as the core of our code and build our \gls{lodegp} on top of their work, which makes it an essential part of our implementation.
Nevertheless, it could be replaced with other \gls{gp} libraries if the kernel would be rewritten for those.
We use SageMath to calculate the symbolic \gls{snf}, given the system of \glspl{ode}.
In our current implementation, SageMath is a necessary library, but could be replaced by any other computer algebra system, with a Python interface, that symbolically solves the \gls{snf} for matrices $A\in \R[x]^{m\times n}$ over the polynomial ring $\R[x]$, and additionally the function field $\R(a_1,\ldots,a_k)[x]$ if there are parameters in the \glspl{ode}.

Training a \gls{lodegp} on 25 multidimensional\footnote{The data is three dimensional for the bipendulum and heating system and five dimensional for the three tank system.} datapoints with the stated training specifications needed, in the median, between 3 and 5 seconds per training and, 7 seconds for the three tank system, due to its greater size, as shown in Table~\ref{tab:runtime}.
This result is in accordance to our findings in Section~\ref{sec:evaluation}, where we discuss that the $\mathcal{O}(n^2)$ calculation takes longer, whereas the $\mathcal{O}(n^3)$ stays roughly the same.
Due to the small number of datapoints, the $\mathcal{O}(n^2)$ calculations have a strong influence on the runtime. 

The \gls{gp} needed roughly 15 seconds per training run.

\begin{table}[h]
    \caption{Median training runtime, in seconds, for 300 iterations on 25 datapoints, for the \gls{lodegp} and the \gls{gp}, for each of the systems. Smaller is better.}
        \centering
    \begin{tabular}{ccc}
    	\toprule
        && training runtime \\
        \midrule
        Bipendulum &\gls{lodegp} & 5.02 \\
                   &\gls{gp}     & 2.34 \\
        \midrule
        Heating    &\gls{lodegp} & 3.12 \\
                   &\gls{gp}     & 2.2  \\
        \midrule
        Three tank &\gls{lodegp} & 7.49 \\
                   &\gls{gp}     & 1.07 \\
        \bottomrule
    \end{tabular}
\label{tab:runtime}
\end{table}

Additionally, generating the model, i.e.\ the \gls{snf} calculation, preparing the differentiated covariance function and initializing the model object, took averagely 0.3 seconds.
Of this time, the \gls{snf} calculation takes, on average, 0.15 seconds.
These calculations are performed only once for the model and are neglegtable, compared to the training runtime.

During development, we encountered an interesting problem we want to document for other developers.
Namely that PyTorch only trains parameters that are part of the model object itself (i.e.\ which can be called using the \verb|self| keyword).
For example, we can't just use a list, that is part of the model object, in which we store the covariance functions, this would cause PyTorch to not recognize the parameters and not train them.
In our case, the various kernels and \gls{ode} parameters, that are added dynamically, required us to give each parameter and kernel a name and add it to the kernel individually.
We also store a position matrix with references to the respective kernel at that position, e.g.\ for the bipendulum covariance function we have a $3\times3$ matrix with different kernel objects for each position.

Additionally, we want to highlight the On-Line Encyclopedia of Integer Sequences, which we used to find the construction formula of the number sequence for the general \gls{se} kernel derivative.\footnote{\url{http://oeis.org/A096713}}

\subsection{Proof for base covariance function construction}\label{app:proof_kernel_construction}
We prove the following Lemma

\begin{namedtheorem}[{Lemma \ref{lemma:base_cov_table}}]
    Covariance functions for solutions of the scalar linear differential equations $d\cdot f = 0$ with constant coefficients, i.e.\ $d\in\R[\partial_t]$, are given by Table~\ref{tab:base_cov_construction} for $d$ is primary, i.e.\ a power of an irreducible real polynomial.
    In the case of a non-primary $d$, each primary factor $d_i$ of $d=\prod_{i=0}^{\ell-1}d_i$ is first translated to its respective covariance function $k_i$ separately before they are added up to give the full covariance function $k(t_1, t_2) = \sum_{i=0}^{\ell-1} k_i(t_1, t_2)$. 
    \begin{table}[h]
    \caption{Primary operators $d$ and their corresponding covariance function $k(t_1, t_2)$.}
        \centering
    \begin{tabular}{cc}
        \toprule
        $d$ & $k(t_1, t_2)$\\
        \toprule
        $1$ & $0$ \\
        $(\partial_t -a)^j$ & $\left(\sum_{i=0}^{j-1} t_1^it_2^i\right)\cdot \exp(a\cdot (t_1+ t_2))$ \\
        $((\partial_t -a - ib)(\partial_t -a + ib))^j$ & $\left(\sum_{i=0}^{j-1} t_1^it_2^i\right)\cdot \exp(a\cdot (t_1 + t_2))\cdot\cos(b\cdot( t_1-t_2))$ \\mid
        $0$ & $\exp(-\frac{1}{2}(t_1-t_2)^2)$ \\
        \bottomrule
    \end{tabular}
    \label{tab:base_cov_construction}
    \end{table}

\end{namedtheorem}

\begin{proof}
    For a function $f$ we have $1\cdot f=0$ if and only if $f=0$.
    Such functions are described by the zero covariance function (and zero mean function).
    
    The real differential equation $(\partial_t -a)^j\cdot f=0$ only has the analytic solutions given symbolically by $\sum_{i=0}^{j-1} a_i\cdot t^i\cdot\exp(a\cdot t)$ for arbitrary $a_i \in\R$.
    This is a finite dimensional space and hence (as for linear regressions problems) a covariance function is given by $\sum_{i=0}^j\left(t_1^i\cdot\exp(a\cdot t_1)\right)\cdot \left(t_2^i\cdot\exp(a\cdot t_2)\right)=\left(\sum_{i=0}^{j-1} t_1^it_2^i\right)\cdot \exp(a\cdot (t_1+ t_2))$.
    
    The real differential equation $((\partial_t-a)^2+b^2)^j=((\partial_t -a - ib)(\partial_t -a + ib))^j$ only has the analytic solutions given symbolically by $\exp(a\cdot t)\cdot\sum_{i=0}^{j-1}t^i\cdot \left(a_i\cdot\cos(b\cdot t)+ b_i\cdot\sin(b\cdot t)\right)$ for arbitrary $a_i\in\R$.
    This is again finite dimensional space and hence a covariance function is given by 
    \begin{align*}
    	&\phantom{=} \sum_{i=0}^j\left(t_1^i\cdot\exp(a\cdot t_1)\cdot\cos(b\cdot t_1)\right)\cdot \left(t_2^i\cdot\exp(a\cdot t_2)\cdot\cos(b\cdot t_2)\right)\\
    	&\phantom{=}+ \sum_{i=0}^j\left(t_1^i\cdot\exp(a\cdot t_1)\cdot\sin(b\cdot t_1)\right)\cdot \left(t_2^i\cdot\exp(a\cdot t_2)\cdot\sin(b\cdot t_2)\right)\\
    	&=\left(\sum_{i=0}^{j-1} t_1^it_2^i\right)\cdot \exp(a\cdot (t_1+ t_2))\cdot\left(\cos(b\cdot t_1)\cdot\cos(b\cdot t_2)+\sin(b\cdot t_1)\cdot \sin(b\cdot t_2)\right) \\
    	&=\left(\sum_{i=0}^{j-1} t_1^it_2^i\right)\cdot \exp(a\cdot (t_1 + t_2))\cdot\cos(b\cdot( t_1-t_2)).
    \end{align*}
    The factor $\cos(b\cdot( t_1-t_2))$ is sometimes called the cosine covariance function.
    
    For a smooth function $f\in C^\infty(\R,\R)$, the equation $0\cdot f=0$ poses no restriction.
    Hence, by \cite[Prop.~1]{langehegermann2022boundary}, the squared exponential covariance function yields a \gls{gp} with realizations densely contained in the space $C^\infty(\R,\R)$.
\end{proof}

\subsection{Licenses and versions}
We use Python 3.9.7, which uses the PSF license agreement.

We use GPyTorch version 1.5.0 for our experiments, GPyTorch is distributed under the MIT license. 

We use SageMath version 9.5, released 2022-01-30, for our experiments, SageMath is distributed under the CC BY-SA 4.0 license.

\end{document}